\documentclass[runningheads]{llncs}
\usepackage{graphicx}
\usepackage{amsmath,amssymb} 
\usepackage{color}
\usepackage[width=122mm,left=12mm,paperwidth=146mm,height=193mm,top=12mm,paperheight=217mm]{geometry}
\usepackage{subfigure}

\usepackage{verbatim}

\newcommand{\degree}{^\circ}

\newcommand{\tabincell}[2]{\begin{tabular}{@{}#1@{}}#2\end{tabular}}
\begin{document}
\pagestyle{headings}
\mainmatter
\def\ECCV18SubNumber{***}  

\title{Invariance Analysis of Saliency Models versus Human Gaze During Scene Free Viewing } 



\author{Zhaohui~Che$^\dag$,
        Ali~Borji$^\ddag$,
        Guangtao~Zhai$^\dag$,
        and~Xiongkuo~Min$^\dag$}
\institute{$^\dag$Shanghai Jiao Tong University (SJTU), $^\ddag$University of Central Florida (UCF)}

\maketitle

\begin{abstract}
Most of current studies on human gaze and saliency modeling have used high-quality stimuli. In real world, however, captured images undergo various types of distortions during the whole acquisition, transmission, and displaying chain. Some distortion types include motion blur, lighting variations and rotation. Despite few efforts, influences of ubiquitous distortions
   on visual attention and saliency models have not been systematically investigated. In this paper, we first create a large-scale database including eye movements of 10 observers over 1900 images degraded by 19 types of distortions. Second, by analyzing eye movements and saliency models, we find that: a) observers look at different locations over distorted versus original images, and b) performances of saliency models are drastically hindered over distorted images, with the maximum performance drop belonging to Rotation and Shearing distortions. Finally, we investigate the effectiveness of different distortions when serving as data augmentation transformations. Experimental results verify that some useful data augmentation transformations which preserve human gaze of reference images can improve deep saliency models against distortions, while some invalid transformations which severely change human gaze will degrade the performance.
\keywords{Human Gaze, Saliency Prediction, Data Augmentation.}
\end{abstract}

\section{Introduction}

Visual attention is an advanced internal mechanism for detecting informative and conspicuous regions from external visual stimuli.
Over digital images or videos, human fixations represent the coordinate value of conspicuous regions are a good proxy of visual attention. Visual attention is an efficient front-end operation for complex back-end computer vision tasks such as scene understanding, object recognition and detection, segmentation and visual description \cite{SalObjDec01,SalObjDec02,SalSegment03}.

A plethora of computational saliency models have been proposed in the past decades
to predict human fixations automatically by simulating human visual system~\cite{borji2013state,borji2013quantitative,Borji_2013_ICCV}. Early saliency models extract hand-crafted features \cite{IttiKoch,GBVS,Torralba,Covsal,AIM,ImageSig,LSLGS,BMS,RC,Murray,AWS,ContextAware}, while deep
saliency models \cite{MLnet,SalGAN,SALICON,SalNet,SAM} learn relevant features automatically. Both types of models generate a scalar-valued saliency map which represents the location and importance of the salient regions.


To the best of our knowledge, most state-of-the-art saliency models and cognitive studies on visual attention have used high-quality and distortion-free stimuli.
However, in most practical circumstances, external stimulus are corrupted by diverse distortions. In addition, some saliency-guided applications like image/video quality assessment \cite{ZhangAliTNN} and object detection and recognition \cite{ObjRecSal}
have to deal with distorted images. Some related works have investigated visual attention with the consideration of distortions. Kim \textit{et al.} \cite{MilanfaSN} investigated visual saliency over noisy images. They found that noise significantly degrades the accuracy of saliency models, and proposed a robust model for noise-corrupted images. Judd \textit{et al.} \cite{JuddLowRes} elaborately investigated the visual fixations on low-resolution images, and compared human gaze dispersions on different resolutions.
Min \textit{et al.} \cite{MinICME} investigated the influences of compression artifacts on visual attention by conducting an eye-tracking experiment on images with different compression levels. Zhang \textit{et al.} \cite{ZhangQA} investigated the optimal strategy to integrate the human attention cues into perceptual quality prediction, and pointed out that eye-tracking data on distorted images promotes perceptual quality metrics' performances.

The above works have considered specific types of distortions
using limited amount of data and a small set of saliency models. In this paper, we conduct a more comprehensive analysis to investigate the influence of distortions on both human gaze and saliency models. We first construct a large saliency database including 1900 images corrupted by 19 types of distortions. Eye-tracking experiments are conducted on this database. We analyze the human gaze discrepancy when viewing stimulus corrupted by different distortions. Then we conduct a comprehensive comparison of several state-of-the-art deep learning based as well as classic saliency models on this database. We measure the gap between the saliency model prediction and human gaze when distortions are introduced. More interesting observations concerning visual attention and stimuli distortions are also made. Our results have important implications for applying saliency models in practical applications with distortions, and provide useful data augmentation strategies to improve deep learning based saliency models.
To the best of our knowledge, this is the first systematic effort in this direction in the saliency field.

\section{The Proposed Eye-tracking Database}
\subsection{Stimuli and distortion types}
We selected 100 distortion-free reference images from the finer-grained \textbf{CAT2000} saliency database \cite{CAT2000} since it covers various scenes including
indoor and outdoor scenes, natural and man-made scenes, synthetic patterns, fractal images, cartoon images, etc.
Specifically, the reference set of the proposed
database consists of 33 outdoor scenes, 22 indoor scenes, 15 cartoon images, 15 art images, and 15 fractal images. Considering that different reference
images have different aspect ratios, the authors of \textbf{CAT2000} padded each image by adding two gray bands to left and right sides and adjusted the image scale to make sure all images have the same resolution ($1080 \times 1920$). This setting guarantees the consistency of the eye-tracking experiment so that eye-movement data will not be affected by resolution.

\begin{table}
\centering
\renewcommand{\arraystretch}{1.6}
\scriptsize
    \caption{ The details of the proposed eye-tracking database. IO score \cite{LSLGS}  provides an upper-bound on prediction accuracy of saliency models.}
    \vspace{-10pt}
    \label{tab:DisGrou}
    \tiny
    \centering
    \begin{tabular}{|c|c|c|}

    \hline
    {\bfseries Distortion Types} &{\bfseries Generation Method (using Matlab)} &{\bfseries IO score : \textcolor{red}{sAUC}, \textcolor{green}{CC}, \textcolor{blue}{NSS} } \\
   \hline
   \hline
    {\emph{\bfseries Reference}} &{\emph{ 100 distortion-free images from CAT2000, denoted by} img}  &{ \textcolor{red}{0.7335}, \textcolor{green}{0.9538}, \textcolor{blue}{3.4352}}\\
    \hline
       \hline
    {{\bfseries MotionBlur1}}  &{{ imfilter(img, fspecial('motion', \bfseries {15}, \bfseries{0}))}} &{ \textcolor{red}{0.6637}, \textcolor{green}{0.9226}, \textcolor{blue}{2.5720}}\\
    \hline
    {{\bfseries MotionBlur2}}  &{{ imfilter(img, fspecial('motion', \bfseries{35}, \bfseries{90}))}} &{ \textcolor{red}{0.6512}, \textcolor{green}{0.9203}, \textcolor{blue}{2.5883}}\\
    \hline
    {{\bfseries Noise1}}  &{{ imnoise(img, 'gaussian', 0, \bfseries{0.1})}} &{ \textcolor{red}{0.7060}, \textcolor{green}{0.9394}, \textcolor{blue}{3.0316}}\\
    \hline
    {{\bfseries Noise2}}  &{{ imnoise(img, 'gaussian', 0, \bfseries{0.2})}} &{ \textcolor{red}{0.6961}, \textcolor{green}{0.9392}, \textcolor{blue}{3.0256}}\\
    \hline
       \hline
    {{\bfseries JPEG1}}  &{{ imwrite(img, saveroutine, 'Quality', \bfseries{5})}} &{ \textcolor{red}{0.7030}, \textcolor{green}{0.9021}, \textcolor{blue}{2.9193}}\\
    \hline
    {{\bfseries JPEG2}}  &{{ imwrite(img, saveroutine, 'Quality', \bfseries{0})}} &{ \textcolor{red}{0.7046}, \textcolor{green}{0.9034}, \textcolor{blue}{2.8633}}\\
    \hline
       \hline
    {{\bfseries Contrast1}}  &{{ imadjust(img, [], \bfseries{[0.3,0.7]})}} &{ \textcolor{red}{0.7220}, \textcolor{green}{0.9306}, \textcolor{blue}{3.0077}}\\
    \hline
    {{\bfseries Contrast2}}  &{{ imadjust(img, [], \bfseries{[0.4,0.6]})}} &{ \textcolor{red}{0.7021}, \textcolor{green}{0.9311}, \textcolor{blue}{3.4303}}\\
    \hline
    {{\bfseries Rotation1}}  &{{ imrotate(img, \bfseries{-45}, 'bilinear', 'loose') }} &{ \textcolor{red}{0.6804}, \textcolor{green}{0.8935}, \textcolor{blue}{2.2865}}\\
    \hline
    {{\bfseries Rotation2}}  &{{ imrotate(img, \bfseries{-135}, 'bilinear', 'loose') }} &{ \textcolor{red}{0.6543}, \textcolor{green}{0.8923}, \textcolor{blue}{2.0978}}\\
    \hline
    {{\bfseries Shearing1}}  &{{ imwarp(img, \bfseries affine2d([1 0 0; 0.5 1 0; 0 0 1]) }} &{ \textcolor{red}{0.7106}, \textcolor{green}{0.9435}, \textcolor{blue}{3.0105}}\\
    \hline
    {{\bfseries Shearing2}}  &{{ imwarp(img, \bfseries affine2d([1 0.5 0; 0 1 0; 0 0 1]) }} &{ \textcolor{red}{0.6874}, \textcolor{green}{0.9273}, \textcolor{blue}{2.5758}}\\
    \hline
    {{\bfseries Shearing3}}  &{{ imwarp(img, \bfseries affine2d([1 0.5 0; 0.5 1 0; 0 0 1]) }} &{ \textcolor{red}{0.6648}, \textcolor{green}{0.8882}, \textcolor{blue}{2.1177}}\\
    \hline
    \hline
    {{\bfseries Inversion}}  &{{ imrotate(img, \bfseries{-180}, 'bilinear', 'loose') }} &{ \textcolor{red}{0.6947}, \textcolor{green}{0.9342}, \textcolor{blue}{3.0621}}\\
    \hline
    {{\bfseries Mirroring}}  &{{ mirror symmetry version of reference images}} &{ \textcolor{red}{0.7256}, \textcolor{green}{0.9306}, \textcolor{blue}{3.3598}}\\
    \hline
    {{\bfseries Boundary}}  &{{ edge(img, 'canny', 0.3, sqrt(2)) }} &{ \textcolor{red}{0.6696}, \textcolor{green}{0.8879}, \textcolor{blue}{2.3119}}\\
    \hline
    {{\bfseries Cropping1}}  &{{ cut a $1080\times200$ narrow band from the \textbf{left side} of img}} &{ \textcolor{red}{0.6972}, \textcolor{green}{0.9343}, \textcolor{blue}{2.6299}}\\
    \hline
    {{\bfseries Cropping2}}  &{{ cut a $200\times1920$ narrow band from the \textbf{top side} of img }} &{ \textcolor{red}{0.6923}, \textcolor{green}{0.9382}, \textcolor{blue}{2.6412}}\\

     \hline
    \end{tabular}
\vspace{-15pt}
\end{table}

To systematically assess the influences of ubiquitous distortions on human and model attention behavior, we choose 19 typical distortions during the whole image acquisition,
 transmission, and displaying chain, explained below : 1) We select 2 levels of motion blur and 2 levels of Gaussian noise to simulate the distortions introduced in the acquisition stage.
2) We consider 2 levels of JPEG compression to simulate distortions introduced in transmission.
3) To simulate displaying distortions, we consider 2 levels of contrast change, 2 rotation degrees, and 3 shearing transformations.
4) In addition, we consider inversion, mirroring, line drawing (sketch/boundary maps), and 2 types of cropping distortions to explore the visual fixation variations under
extremely abnormal conditions. This way, we derive 18 distorted images for each reference stimuli. Thus, a total of 1900 images (18 $\times$ 100 + 100 reference images) are included in the proposed eye-tracking database\footnote{Download link: Obscured for blind review.}. Details of distortion types and generation methods are shown in Table \ref{tab:DisGrou}.

\subsection{Eye tracking apparatus}
Collecting eye-tracking data is expensive and time consuming. To overcome this  challenge, some new large-scale data collection methodologies \cite{SALICONDB,DB2,DB3} have resorted to mouse movements or webcam gaze tracking. These methodologies are devised for reducing the semantic gap between human and deep models. However, it is still unclear whether they can replace eye movements for model training and whether they suffice to reach human level accuracy. Therefore, instead of relying on such methods, here we employ laboratory eye tracking which is more accurate.

Bylinskii \textit{et al.} \cite{Bylinskii} pointed out that the eye-tracking parameters, such as the distance of the subject to eye-tracker, calibration error, and image size affect quality of the collected data. Poor experimental settings can significantly affect
performance evaluation and conclusions. Here we utilize \textbf{Tobii X120} eye tracker to record eye-movements. 
We use the \textbf{LG 47LA6600 CA} monitor with horizontal resolution of 1920 and vertical resolution
 of 1080, so that the resolutions of stimulus and the monitor screen are the same. The height and width of the monitor are 60cm and 106cm, respectively. 
 The distance between the subject
  and the monitor is 180cm, and the distance between the subject and the eye-tracker is 60cm. According to Bylinskii \textit{et al.} \cite{VisualAngle}, one degree of visual angle is used both as
  1) an estimate of the size of the human fovea: e.g., how much of the image a participant has in focus during a fixation, and
  2) to account for measurement error in the eye-tracking set-up. In our experiment, the width of the screen subtends $32.81\degree$ of visual angle,
  and $1\degree$ of horizontal visual angle contains 56.91 pixels. Accordingly, the height of the screen subtends $18.92\degree$ of visual angle, and $1\degree$
  of vertical visual angle corresponds to 56.55 pixels. The obtained visual angle is necessary for computing the standard deviation of Gaussian kernel used in the following step.

  It is well known that there are two types of ground-truths for measuring the performances of visual saliency models, i.e.
   1) a discrete fixations map made up of discrete gaze points which are recorded by an eye-tracker directly, and
   2) a continuous fixation map representing the probability of the human fixation. The former can be converted into the latter by a Gaussian smooth filter with
   standard deviation $\sigma$ equal to one degree of visual angle \cite{LeMeur2013}. Here, we choose $\sigma=57$. 

\subsection{Subjects and task}
We recruited 40 subjects to participate in the eye tracking experiment. They were 24 males and 16 females, with age ranging from 18 to 35 years old. All participants were naive subjects and had not seen the stimulus set before. Besides, subjects viewed the stimulus under a free-viewing condition.

Engelke \textit{et al.}\cite{FinallyFind} investigated the impact of eye-tracking experimental settings on the quality of fixation maps, and found that the fixation map
becomes more stable with the longer duration time. Further, they pointed out that the convergence speed of fixation map may aid in reducing experimental time and cost while marginally sacrificing
the accuracy of the final fixation map. Moreover, for duration times longer than $t = 4s$, the fixation map is accurate enough to approximate the human attention behavior. In the proposed database, the duration time for each stimuli is $4s$. We inserted a gray image with $t = 1s$ duration time between each two consecutive images to reset the visual fixation center and avoid carryover and memory effects \cite{WithinSubject}. Besides, the order of stimulus was randomized for each subject to mitigate the carryover effect.

\subsection{``Between-subjects'' protocol}

A traditional subjective experiment protocol called ``within-subjects'' has been widely utilized in eye-tracking experiments
\cite{MinICME,AnotherMin}. The ``within-subjects'' protocol asks the same group of subjects to view all stimulus, so that each subject has to
view several distorted versions originated from the same reference image. This protocol may cause some undesirable effects such as learning from experience or memory about salient objects and the carryover effect \cite{WithinSubject}. Hence, we adopt the ``between-subjects'' experiment protocol proposed in \cite{BetweenSubjects} instead of the ``within-subjects'' protocol. In ``between-subject'' protocol, the subjects are divided into several non-overlapping groups. Each group is randomly assigned to view different distortion groups. This protocol reduces the repetitive stimulus presented to each single subject.
Considering that the proposed database
 contains 1 reference group and 18 distortion groups, we equally divided the 40 subjects into 4 groups, and arranged where each subject
 viewed only 4 or 5 distortion groups, rather than all 19 groups. As a result, the carryover effect will be mitigated. This way, for each stimulus, we collect eye-movement data from 10
 subjects. For each subject, it takes 42 minutes to accomplish the eye-tracking experiment.

\section{Analysis of Human Gaze Discrepancy}
\subsection{Quantitative evaluation}
In this section, we investigate the human visual fixation dispersions over 19 distortions, and compare the fixation maps of the reference and distorted images. 

\begin{figure*}
\subfigure[\scriptsize CC similarity matrix]{\label{fig:edge-a}\includegraphics[height=0.36\linewidth]{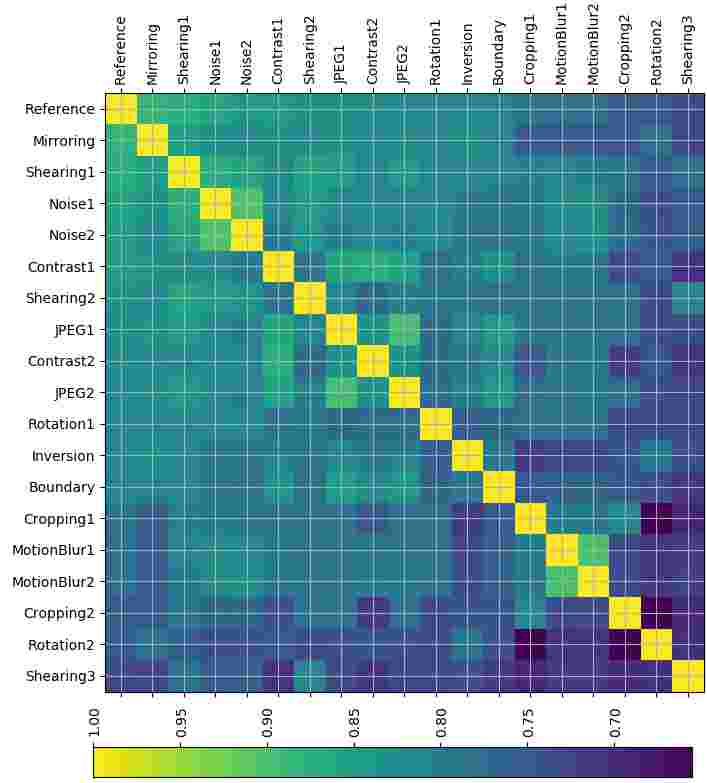}}
\subfigure[\scriptsize SIM similarity matrix]{\label{fig:edge-a}\includegraphics[height=0.36\linewidth]{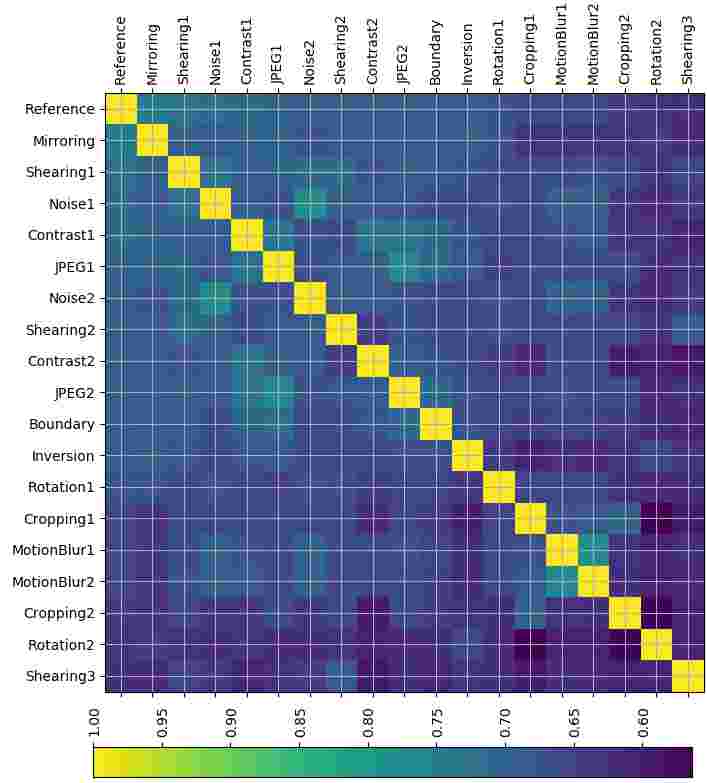}}
\subfigure[\scriptsize KL dissimilarity matrix]{\label{fig:edge-a}\includegraphics[height=0.36\linewidth]{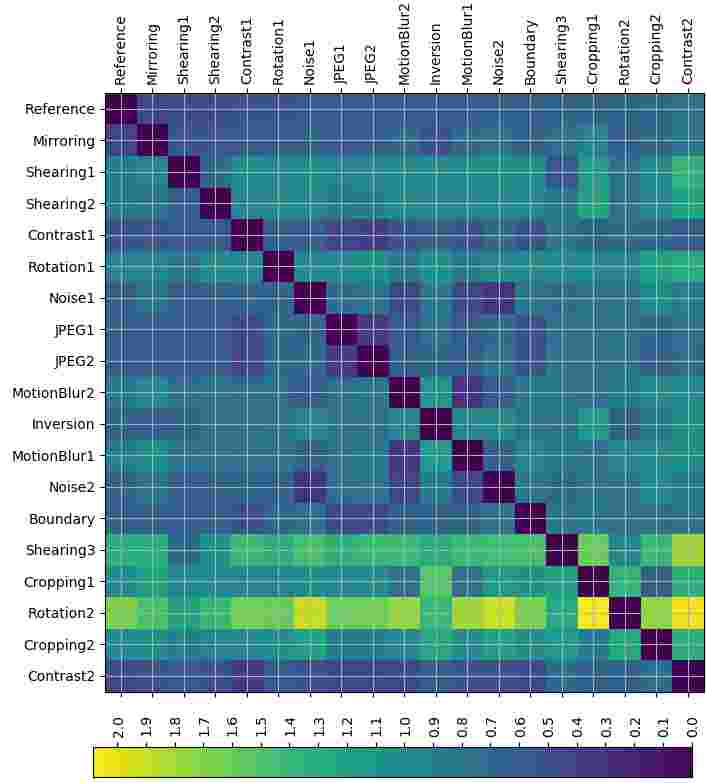}}\\
\vspace{-0.1cm}
\vspace{-0.1cm}
\vspace{-10pt}
\caption{\small CC, SIM similarity matrixes and KL dissimilarity matrix of human gaze on different distortions. The distortion types are ranked by their similarity/dissimilarity values when using the human gaze on Reference as ground-truth. The higher CC and SIM values represent the better similarity, while the lower KL value means the better relevance.}
\label{SMhuman}
\end{figure*}

We quantify the discrepancies between human fixation maps of distorted and reference images using 3 similarity evaluation measures including: Correlation Coefficient (CC), Similarity Measure (SIM), and Kullback-Leibler divergence (KL) \cite{mit-saliency-benchmark}. The similarity matrices of different distortions are shown in Figure \ref{SMhuman}. The distortion types of each similarity matrix are ranked by its similarity value with the reference group, and the relevances decrease from left to right. Figure \ref{SMhuman} indicates that different distortions do have influences on human attention, and the extent of impact is highly related to distortion types. Notably, for Inversion, Mirroring, Rotation and Shearing distortions which change the image pixels' locations,
we map the human gaze maps via the inverse transformations corresponding to Table \ref{tab:DisGrou} to align them with the Reference image pixel-by-pixel for fair comparison, as shown in Figure \ref{R2S3}.
We find that:

\textbf{1.} Mirroring and Shearing1 have slight influences on human attention compared to other distortions, because they obtain the best similarity values in terms of CC, SIM and KL metrics.

\textbf{2.} Rotation2, Cropping2 and Shearing3 have significant influences on human gaze, because the discrepancies of human fixation between these distortions and Reference are significant, as shown by CC, SIM and KL metrics in Figure \ref{SMhuman}.

\textbf{3.} The human fixation maps of the images degraded by the same type but different levels of distortion are quite close for these cases: Noise1 \emph{vs} Noise2, JPEG1 \emph{vs} JPEG2, MotionBlur1 \emph{vs} MotionBlur2 (when compared to each other). Human gaze maps over these distortions in different levels achieve high similarity values using CC, SIM and KL metrics. Besides, the higher distortion level, the higher human gaze discrepancy (when compared to Reference).

\subsection{Finer-grained analyses}
\subsubsection{Characteristics of different metrics:}
As shown in Figure \ref{SMhuman}, different similarity matrices have disparate characteristics including symmetry properties and distortion rankings due to properties of different metrics. A finer-grained analyses about the influence of different metrics including CC, SIM and KL to each distortion type is provided in the supplementary material.

\begin{figure}
\centering
\vspace{-0.1cm}
\tiny
\subfigure[\scriptsize Rotation2]{\label{fig:edge-a}\includegraphics[height=0.32\linewidth]{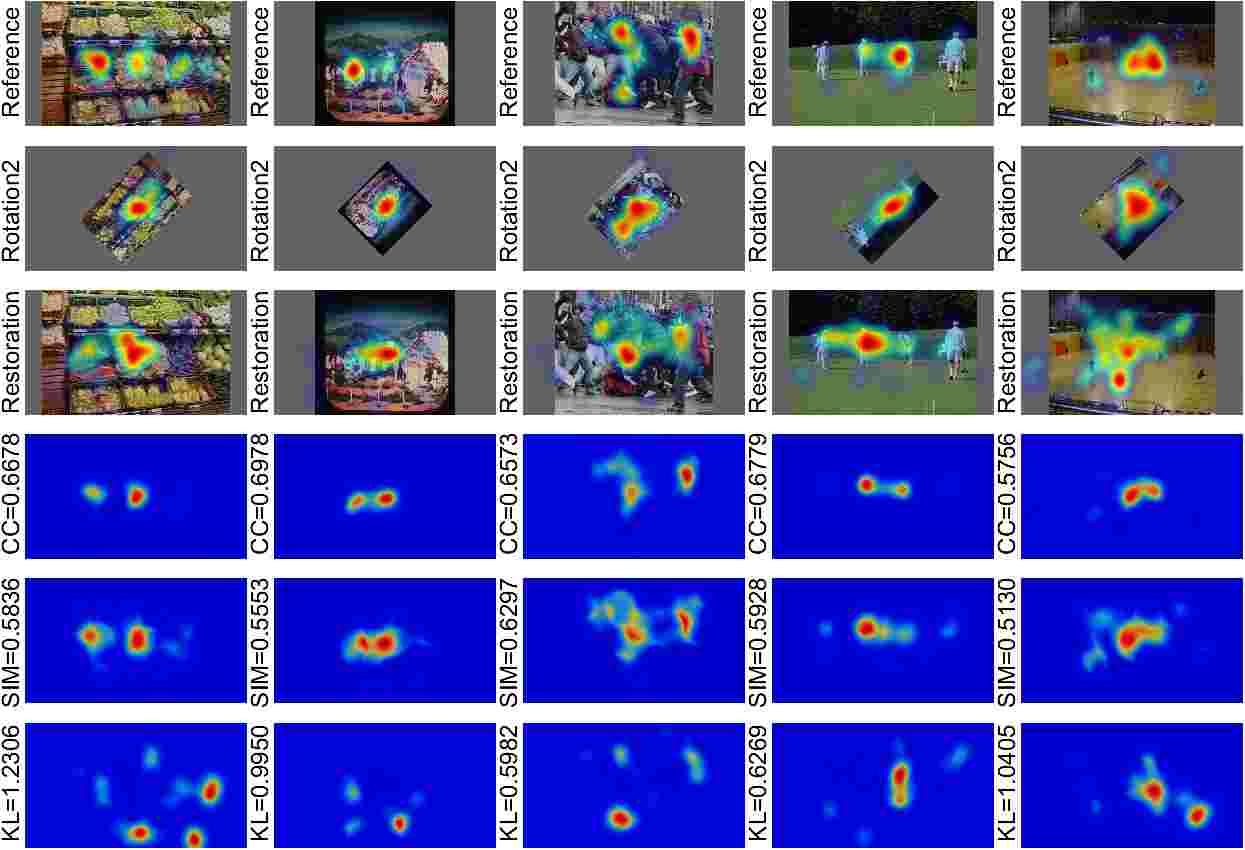}}\hspace{0.2cm}
\subfigure[\scriptsize Shearing3]{\label{fig:edge-a}\includegraphics[height=0.32\linewidth]{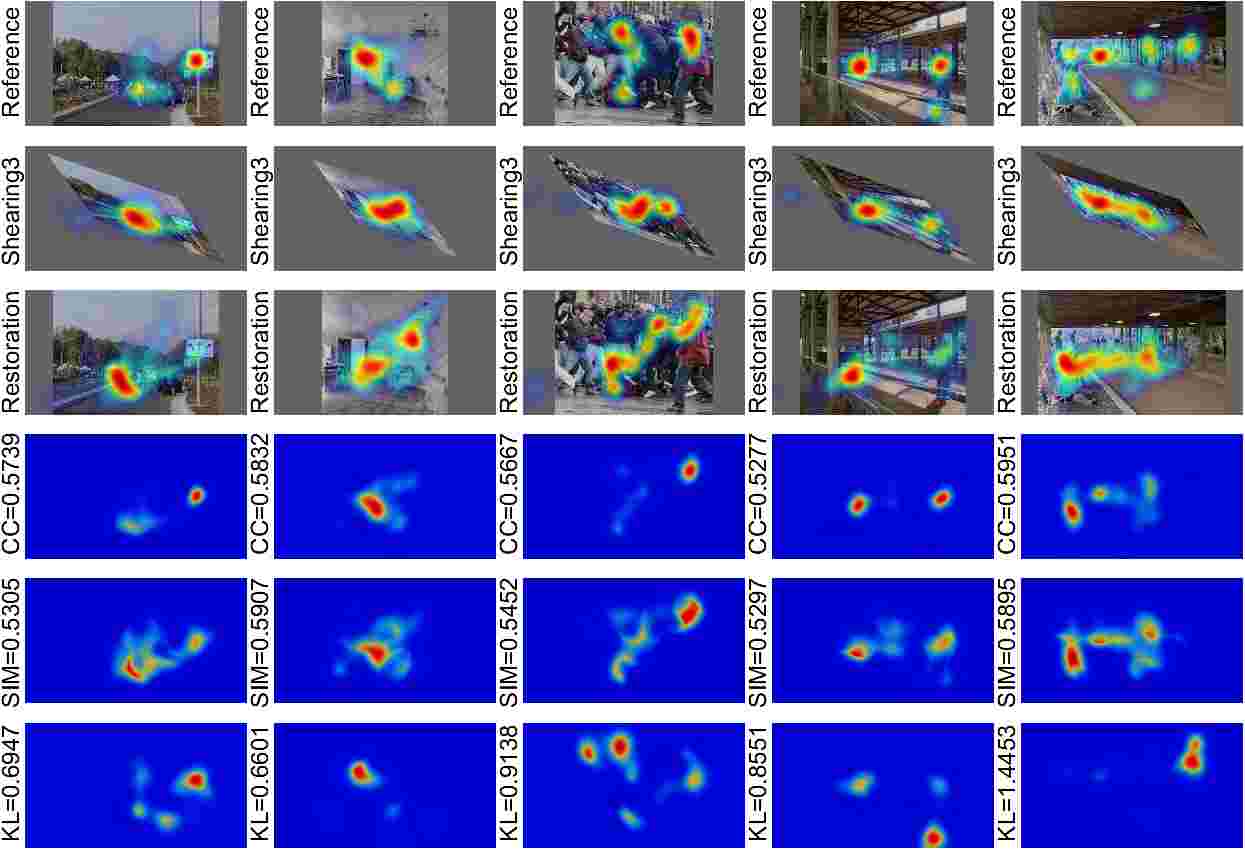}}\\
\vspace{-10pt}
\caption{\small Human gaze discrepancy on Rotation2 and Shearing3 compared to Reference. The 3rd row represents the restored version of Rotation2/Shearing3 via inverse transformation. This way, the Restoration is aligned with Reference pixel-by-pixel for fair comparison. The 1st and 2nd rows represent the human gaze maps of Reference and Cropping1/Cropping2 respectively. The 3rd and 4th rows represent CC and SIM maps in which the higher value means the better approximation. The 5th row represents KL map in which the higher value means the severer discrepancy.}
\label{R2S3}
\end{figure}

\begin{figure}
\centering
\vspace{-0.1cm}
\tiny
\subfigure[\scriptsize Mirroring]{\label{fig:edge-a}\includegraphics[height=0.26\linewidth]{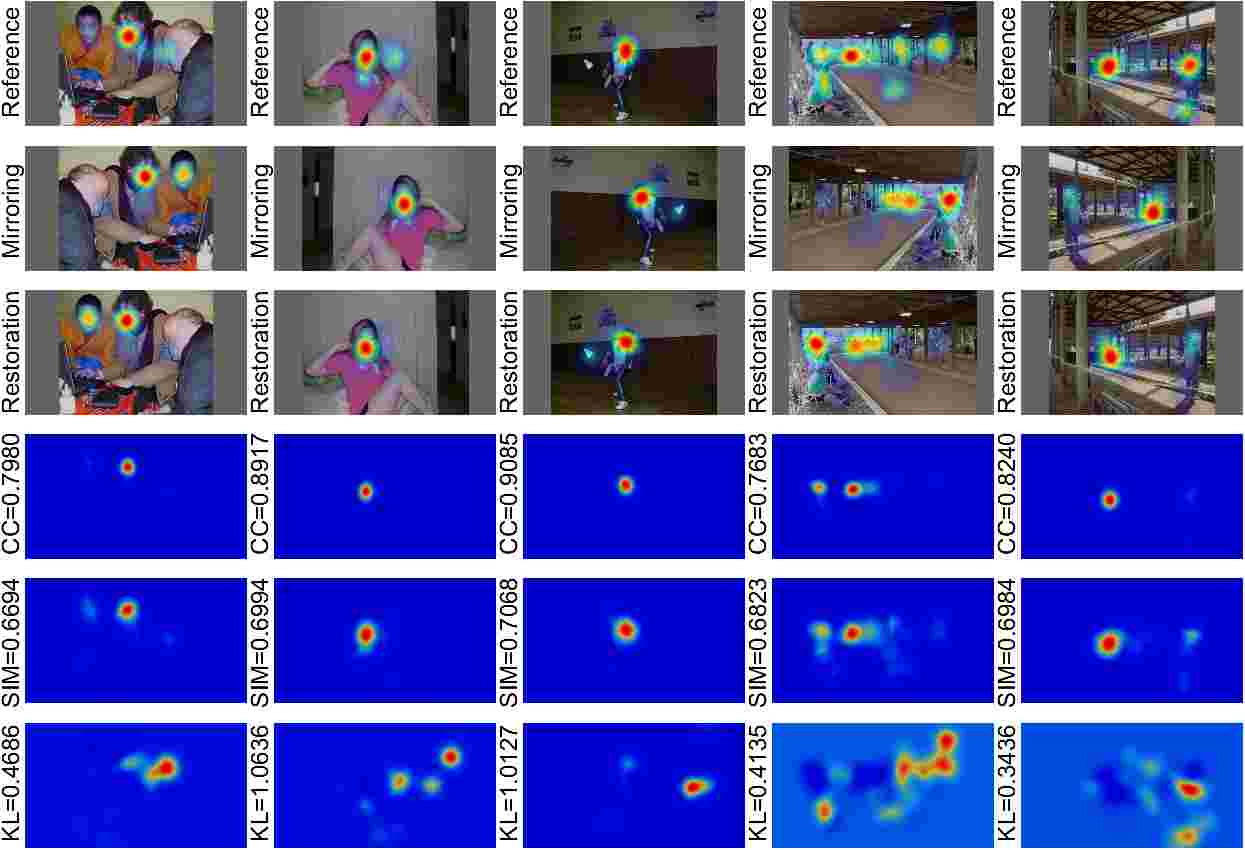}}\hspace{0.2cm}
\subfigure[\scriptsize Boundary]{\label{fig:edge-a}\includegraphics[height=0.26\linewidth]{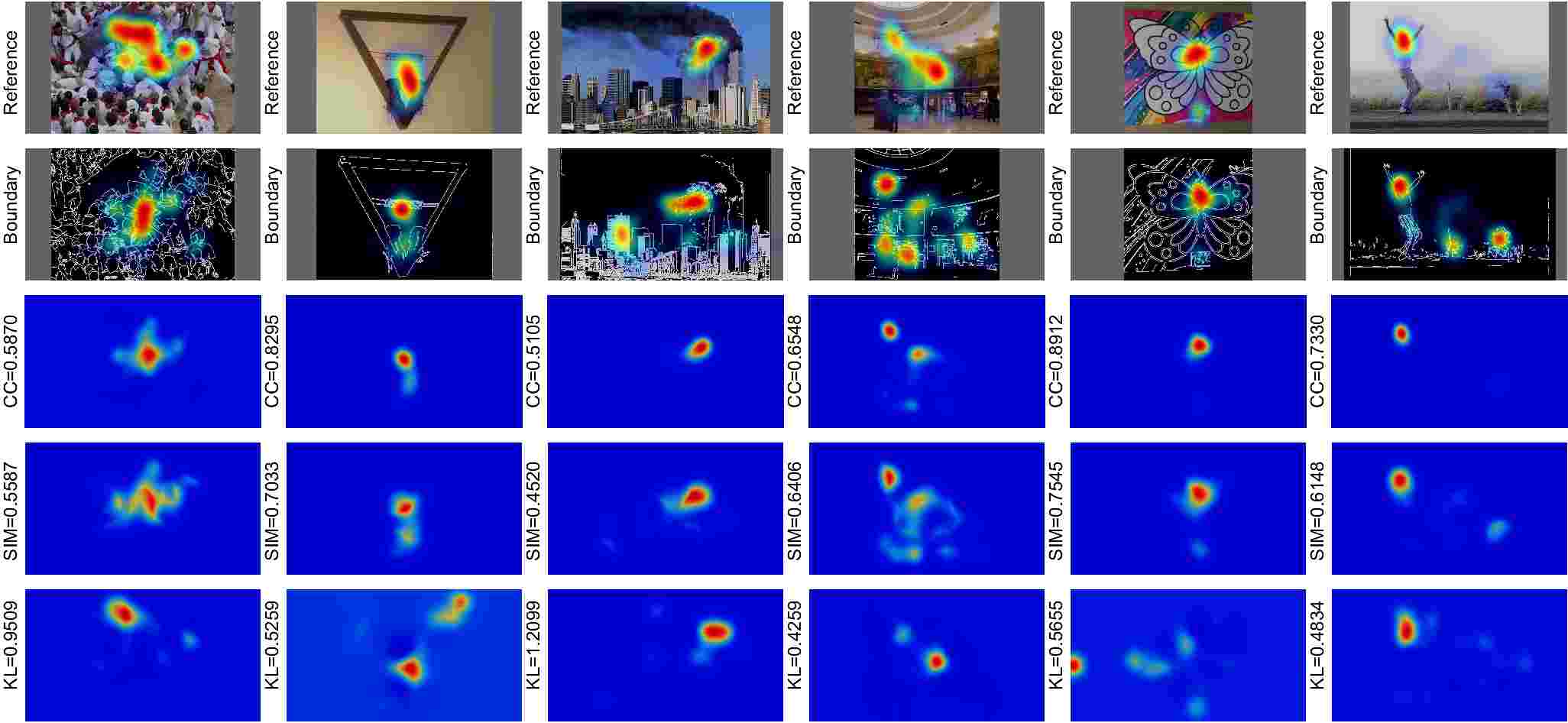}}\\
\vspace{-10pt}
\caption{\small Human gaze discrepancy on Mirroring and Boundary  compared to Reference.}
\label{MiBo}
\end{figure}

\begin{figure}
\centering
\vspace{-0.1cm}
\tiny
\subfigure[\scriptsize Cropping1]{\label{fig:edge-a}\includegraphics[height=0.26\linewidth]{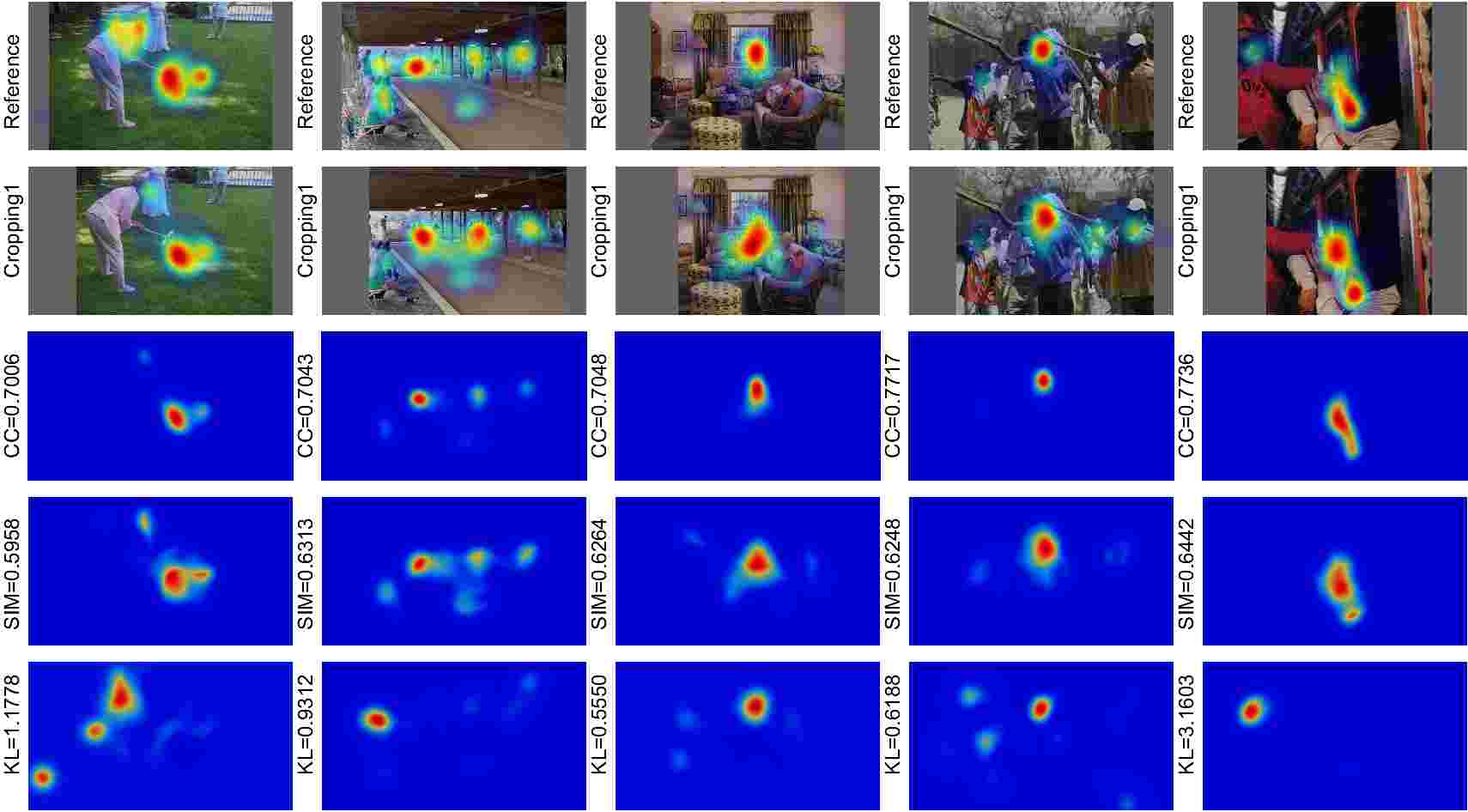}}\hspace{0.2cm}
\subfigure[\scriptsize Cropping2]{\label{fig:edge-a}\includegraphics[height=0.26\linewidth]{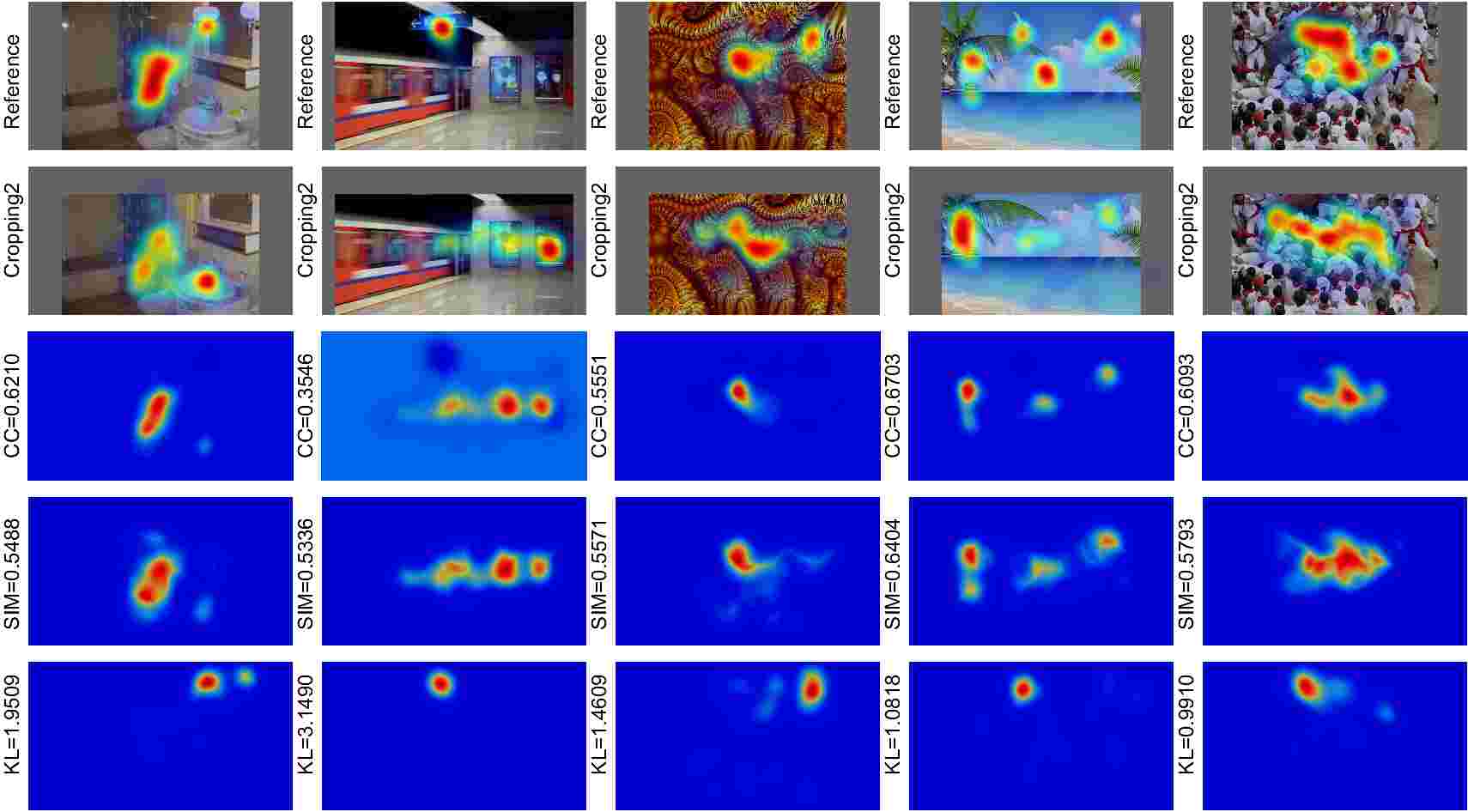}}\\
\vspace{-10pt}
\caption{\small Human gaze discrepancy on Cropping1 and Cropping2  compared to Reference. }
\label{CP1CP2}
\end{figure}

\subsubsection{Influences of different distortions:}
In this section, we summarize the influence of different distortions on human gaze.  

$\mathbf{Mirroring}$: Human gaze maps are almost the same on Mirroring and Reference groups, but there is still a small gap. Generally speaking, for most stimulus with multiple salient objects, the most conspicuous salient object will be noticed in both of Reference and Mirrored images. However, for secondary salient objects, the human fixations may be different on Reference and Mirrored images, as shown in Figure \ref{MiBo}.(a).


$\mathbf{Boundary}$: In general, Boundary group retains most semantic information compared to Reference because the human gaze discrepancy between
Boundary and Reference groups is not huge, even better than Cropping1. We find that humans prefer to look at regions with intensive edges when color and luminance features are lacking, as shown in Figure \ref{MiBo}.(b).

$\mathbf{Cropping}$: Cropping1 distracts human attention from the salient regions appearing on the left side of the whole stimuli, but the main part of the stimuli will not be influenced severely, as shown in Figure \ref{CP1CP2}.(a). Cropping2 alters the human gaze severely, because salient objects containing semantic information are often framed in the center part. As a result, the risk of damaging the objects with semantic information is higher for Cropping2 compared to Cropping1.

$\mathbf{Rotation,Inversion}$: Inversion is a special rotation with $180\degree$ rotation angle. Rotation distortions do have influences on human gaze. Rotation2 with $135\degree$ rotation angle has the severer influence on human attention compared to $45\degree$ (Rotation1) and $180\degree$ (Inversion). In particular, for stimuli with multiple salient objects in a complex background, humans prefer to concentrate on one of the salient objects, and the dominant salient object may be altered, as shown in the 1st, 2nd, and 4th columns of Figure \ref{R2S3}.(a).

$\mathbf{Shearing}$: As shown in Figure \ref{R2S3}.(b), shearing distortions have influences on human gaze, and the strength of influence highly depends on the affine transformation matrix shown in Table \ref{tab:DisGrou}. The severer deformation increases the discrepancy of human gaze when compared to Reference. Considering that geometric distortions (i.e., Rotation and Shearing) change the effective size of images (as shown in Figure \ref{R2S3}), we take some arrangements to mitigate the additional influence of image effective size to eye movement data, and the arrangements are explained in detail in the supplementary material.

$\mathbf{Contrast}$: The low level Contrast1 has slight influence on human gaze, but the high level Contrast2 attracts human gaze to center region, i.e., there is severe center-bias. The qualitative results are provided in the supplementary material.

$\mathbf{Noise,JPEG,MotionBlur}$: Experimental results show that human gaze is tolerant to Gaussian noise maybe because human eyes are able to detect salient regions even when the stimuli is corrupted by severe noise. However, for stimuli including one dominant salient object in a complex background, high level noise will distract human attention from the dominant salient object. Similarly, low level JPEG artifacts will be ignored for human attention, but high level JPEG artifacts will alter the human gaze. MotionBlur artifacts have a more profound impact on human gaze compared to JPEG and Noise.

\begin{figure*}
\centering
\subfigure[\scriptsize CC scores ($std_m=0.0862$,$std_d=0.0263$)]{\label{fig:edge-a}\includegraphics[height=0.4\linewidth]{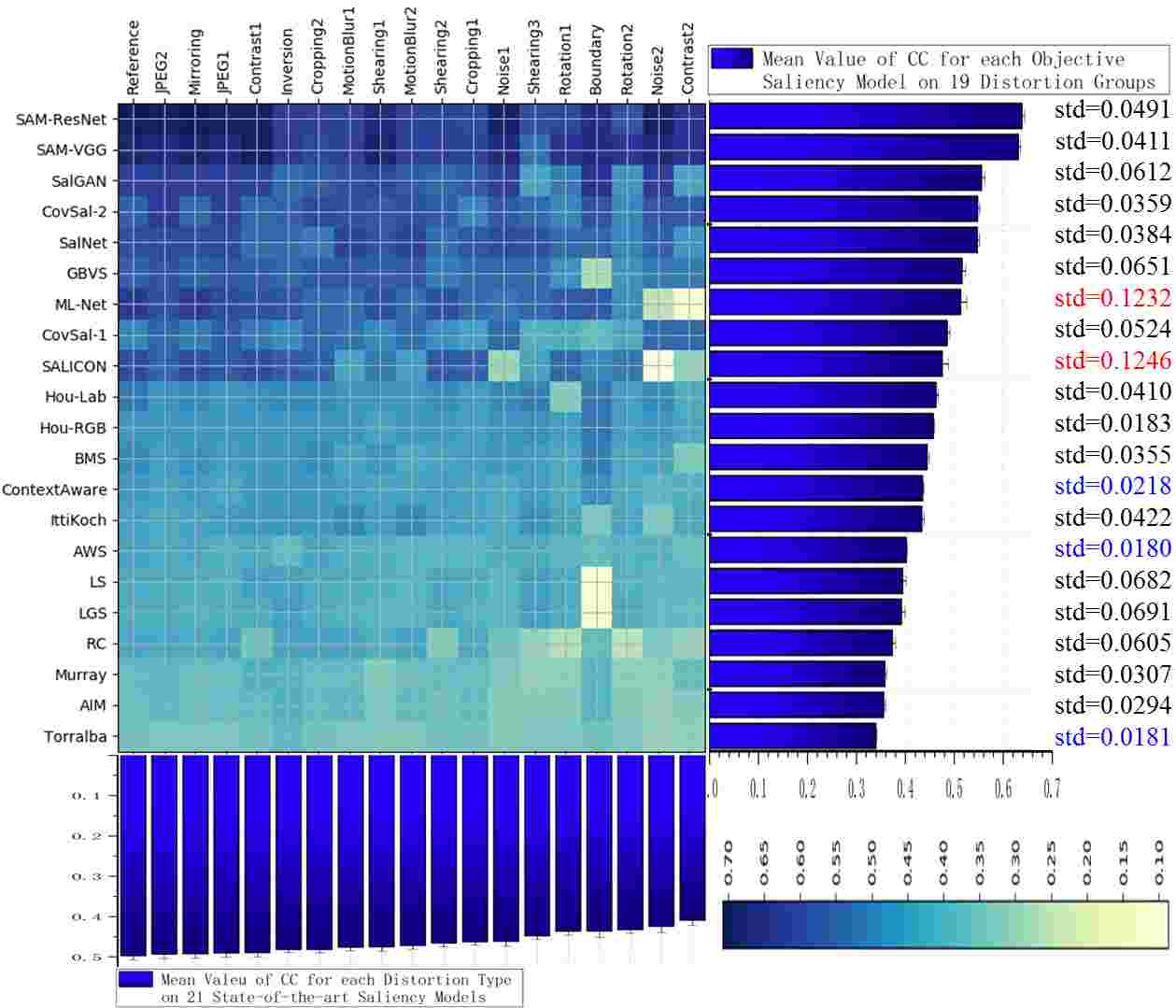}}
\subfigure[\scriptsize NSS scores ($std_m=0.0809$,$std_d=0.0209$)]{\label{fig:edge-a}\includegraphics[height=0.4\linewidth]{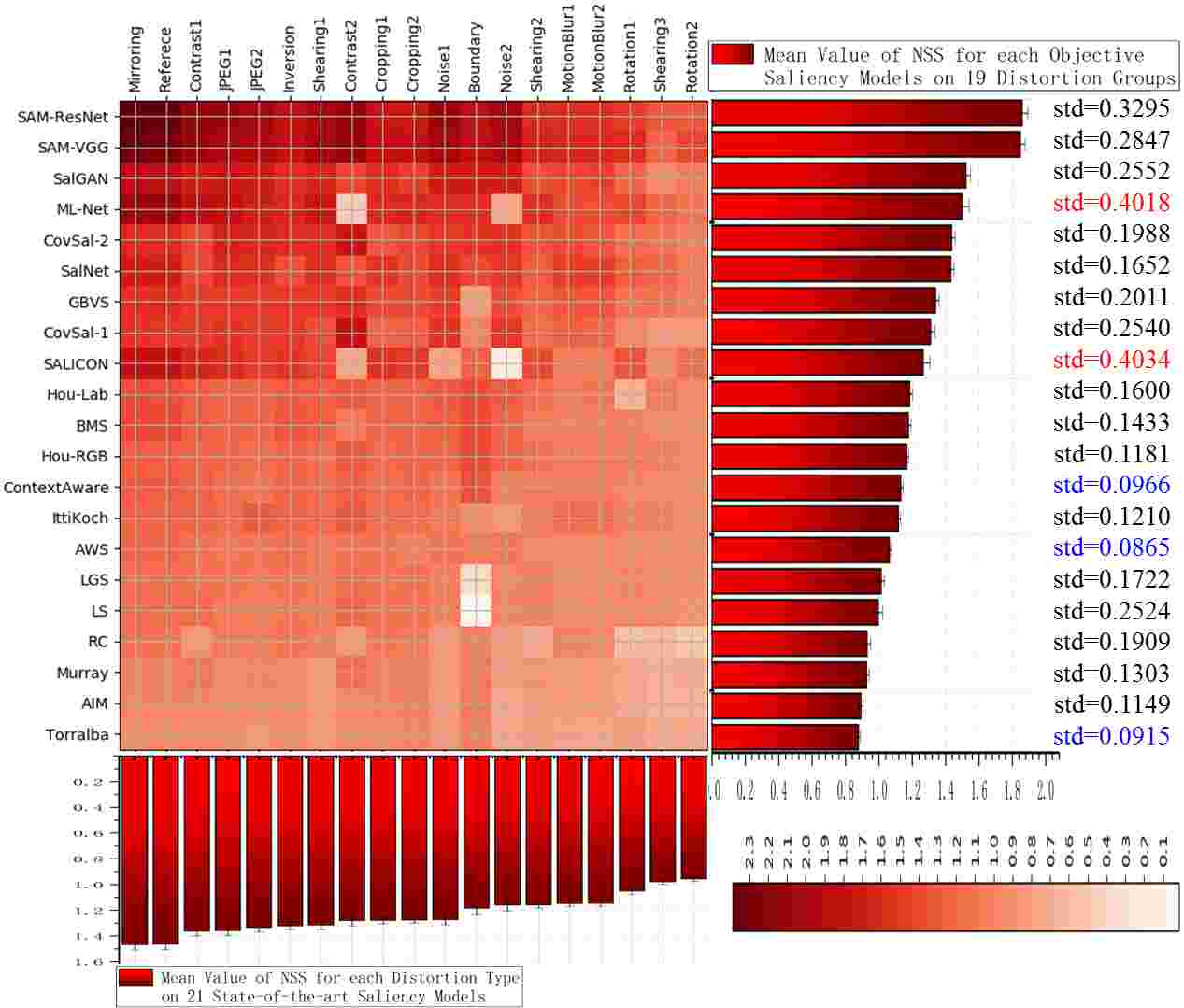}}
\subfigure[\scriptsize sAUC scores ($std_m=0.0356$,$std_d=0.0238$)]{\label{fig:edge-a}\includegraphics[height=0.4\linewidth]{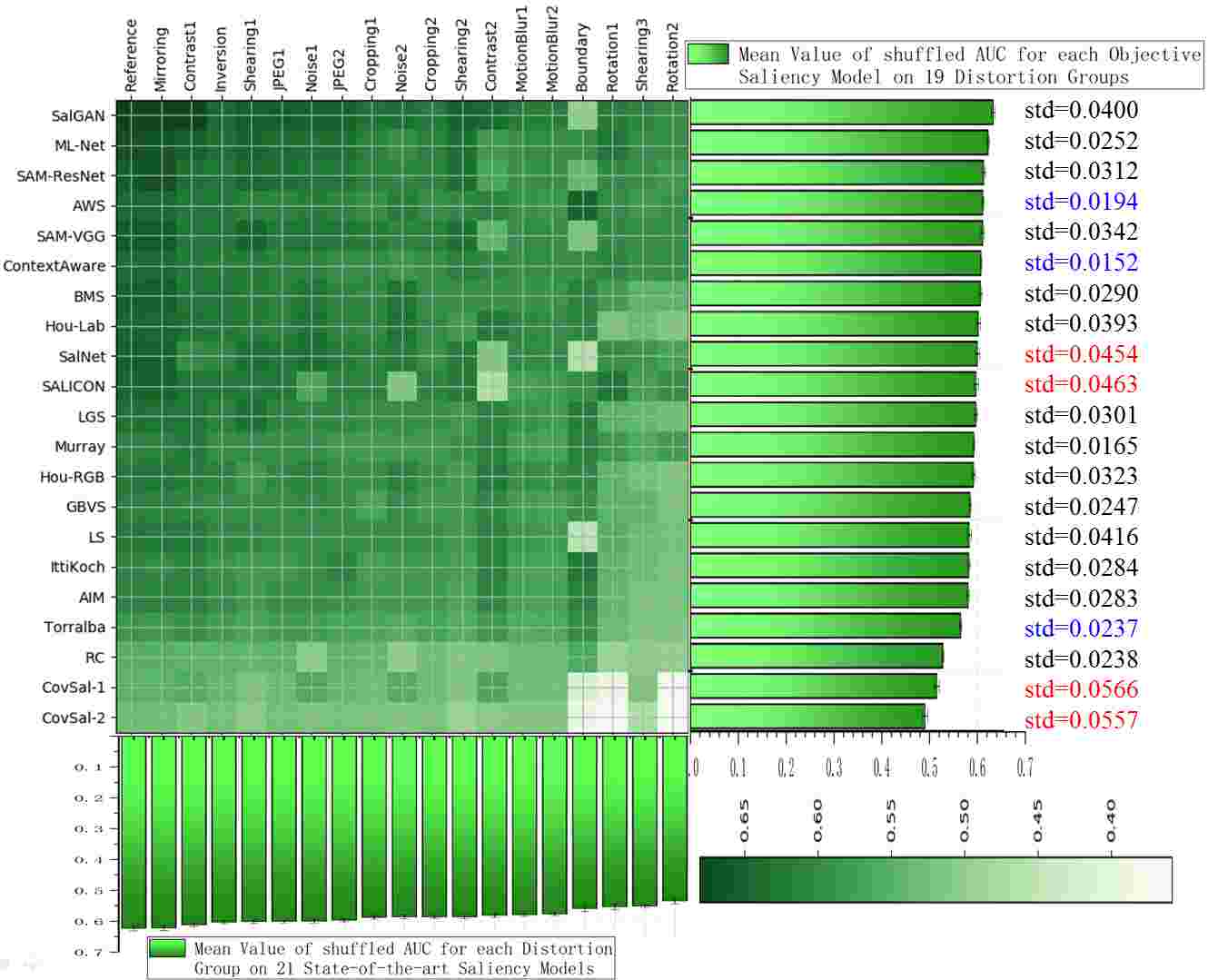}}
\subfigure[\scriptsize sAUC scores when using Human Gaze map on Reference as ground-truth]{\label{fig:edge-a}\includegraphics[height=0.4\linewidth]{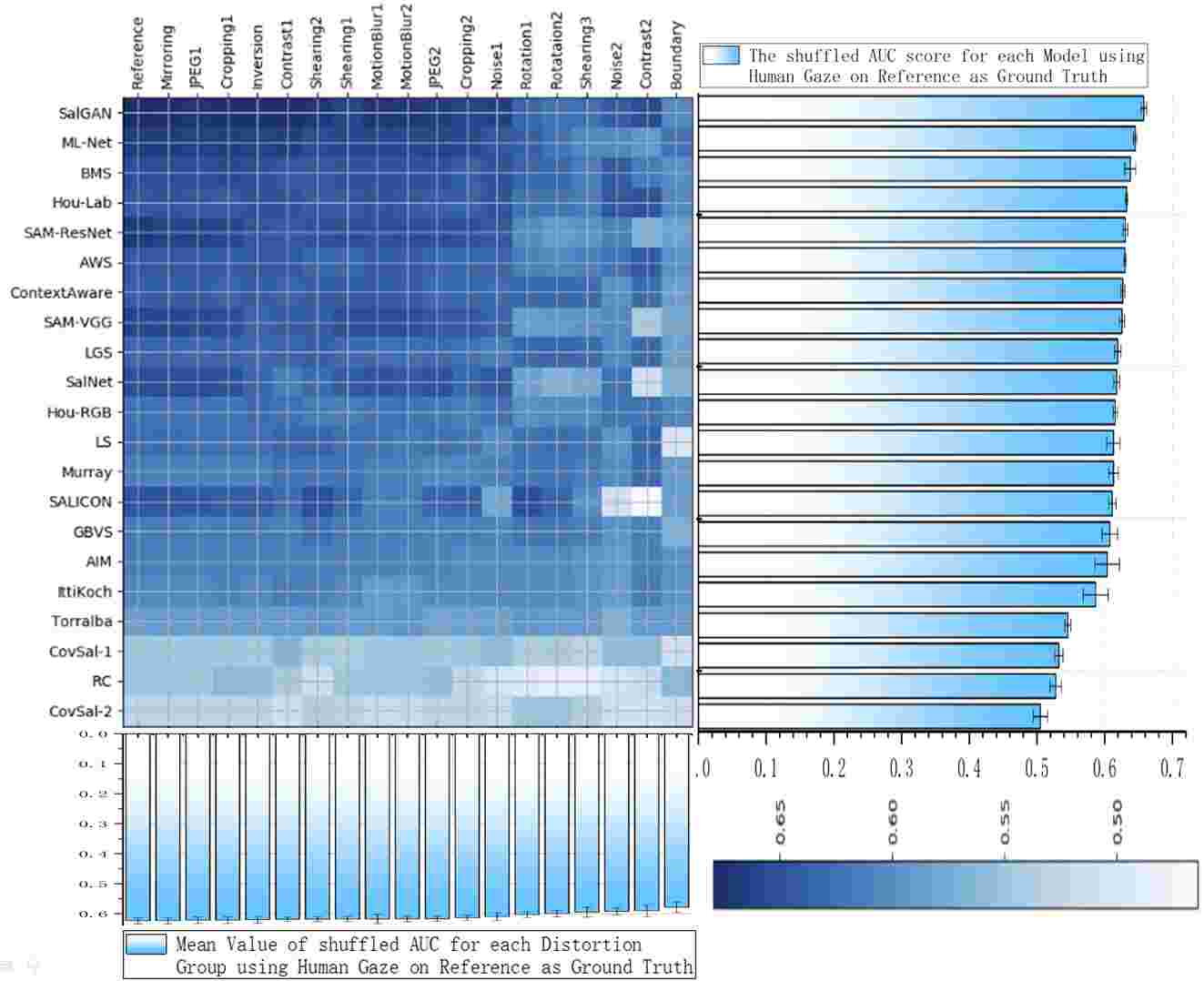}}\\
\vspace{-5pt}
\caption{\small The performance of state-of-the-art saliency models on different distortions. The horizontal axis represents different distortion types which are ranked by average performance over 21 saliency models. The vertical axis represents different saliency models which are ranked by average performance over 19 distortions. The bar graph on the right side represents the average performance of each model, and the bar graph at the bottom represents the average performance of each distortion. The error bars represent standard error of the mean (SEM). The $std$ value means the standard deviation of each model's performance over 19 distortions, and the red and blue $std$ represent the highest and the lowest $std$ values respectively. Notably, $std_m$ represents the standard deviation of the average performances of different models, while $std_d$ represents the standard deviation of the average performances of different distortions. (d): Compared to (c), we calculate the sAUC score for each model once again. However, we adopt the human gaze map of the reference stimuli to compute the sAUC scores for the other 18 distorted stimulus, rather than using the human gaze map of the real distorted stimulus as (a)-(c) do. }
\label{ModelPer}
\end{figure*}
\section{Performance of Saliency Models}
\subsection{Quantitative evaluation}
We test 15 early saliency models including IttiKoch \cite{IttiKoch}, GBVS \cite{GBVS}, Torralba \cite{Torralba}, CovSal \cite{Covsal} (CovSal-1 utilizes covariance feature and CovSal-2 utilizes both of covariance and mean features), AIM \cite{AIM}, Hou \cite{ImageSig} (Hou-Lab and Hou-RGB adopt Lab and RGB color spaces respectively), LS \cite{LSLGS}, LGS \cite{LSLGS}, BMS \cite{BMS}, RC \cite{RC}, Murray \cite{Murray}, AWS \cite{AWS} and ContextAware \cite{ContextAware}, and 6 deep models including ML-Net \cite{MLnet}, SalGAN \cite{SalGAN}, SALICON \cite{SALICON}, SalNet \cite{SalNet}, SAM-ResNet \cite{SAM} and SAM-VGG \cite{SAM} on the proposed database. The performances are shown in Figure \ref{ModelPer}. 

We observe the following points:

\textbf{1.} Deep models outperform early models significantly on different distortions. 

\textbf{2.} Rotation2 and Shearing3 are the most challenging distortions for models, because most models obtain poor performances on these distortions. Recall that Rotation2 and Shearing3 also have severe impacts on human gaze.

\textbf{3.} The discrepancy between different saliency models seems much larger than the discrepancy between different distortions. As shown in Figure \ref{ModelPer}, the standard deviation of average performances of 21 saliency models (i.e., $std_m$) is higher than the standard deviation of average performances of 19 distortions (i.e., $std_d$) when using both of sAUC, CC and NSS metrics.

\textbf{4.} AWS, ContextAware and Torralba models are robust to different distortions, because sAUC, CC and NSS scores of these models obtain small standard deviations (i.e. $std$) over 19 distortions. However, SALICON and ML-Net models obtain unstable performance on CC and NSS metrics, because they fail on Noise2 and Contrast2. The same observation holds for the SalNet and SALICON when using the sAUC metric.
We find that the early models using hand-crafted features obtain more robust performance compared to deep models, while deep models have the higher average performance compared to the early models.

\subsection{Finer-grained analyses}
In this section, we will explain some outliers appearing on Figure \ref{ModelPer}, and explore the gap between saliency models and human gaze.

\begin{figure}
\centering
\vspace{-0.1cm}
\tiny
\subfigure[\scriptsize Boundary]{\label{fig:edge-a}\includegraphics[height=0.26\linewidth]{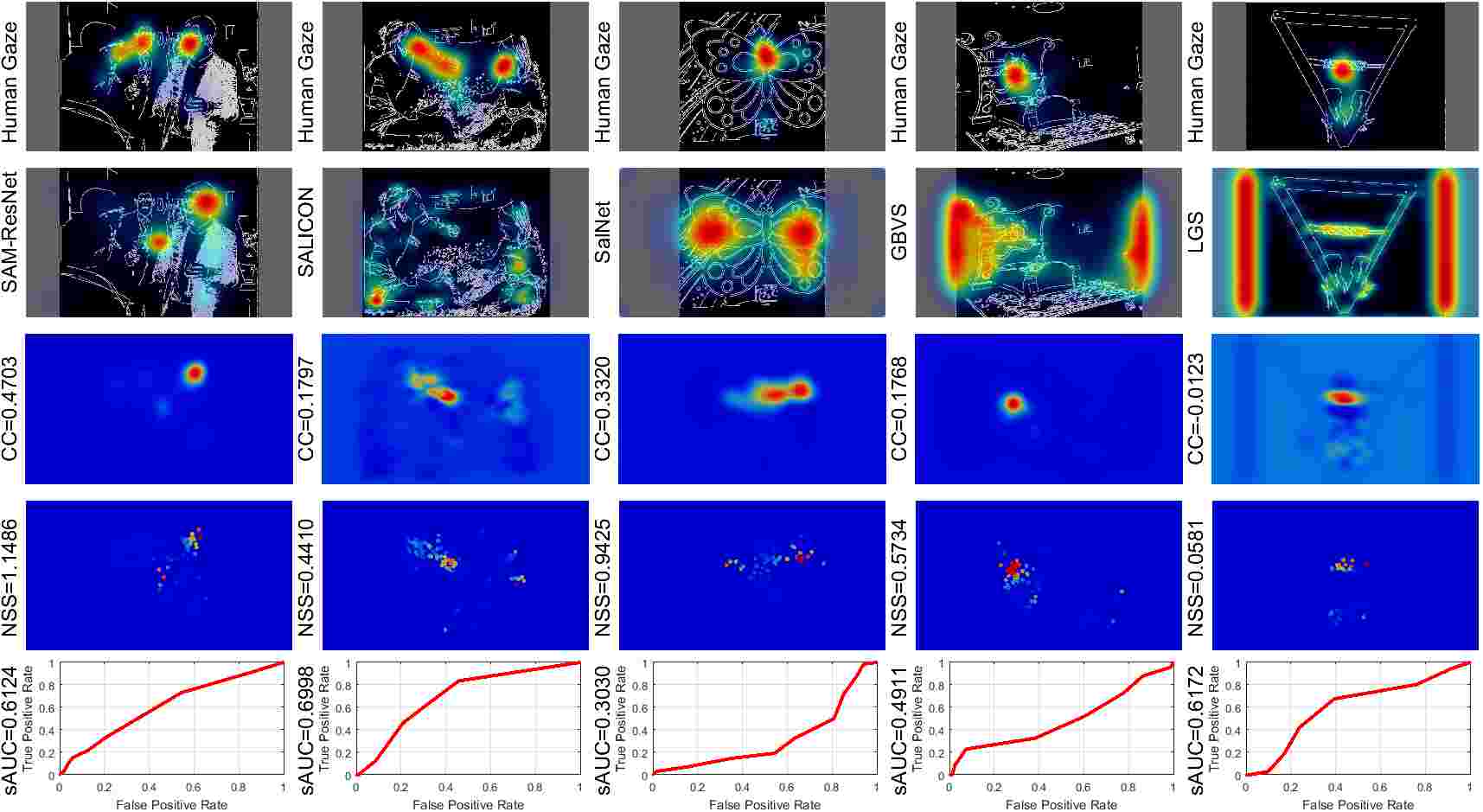}}\hspace{0.2cm}
\subfigure[\scriptsize Rotation2]{\label{fig:edge-a}\includegraphics[height=0.26\linewidth]{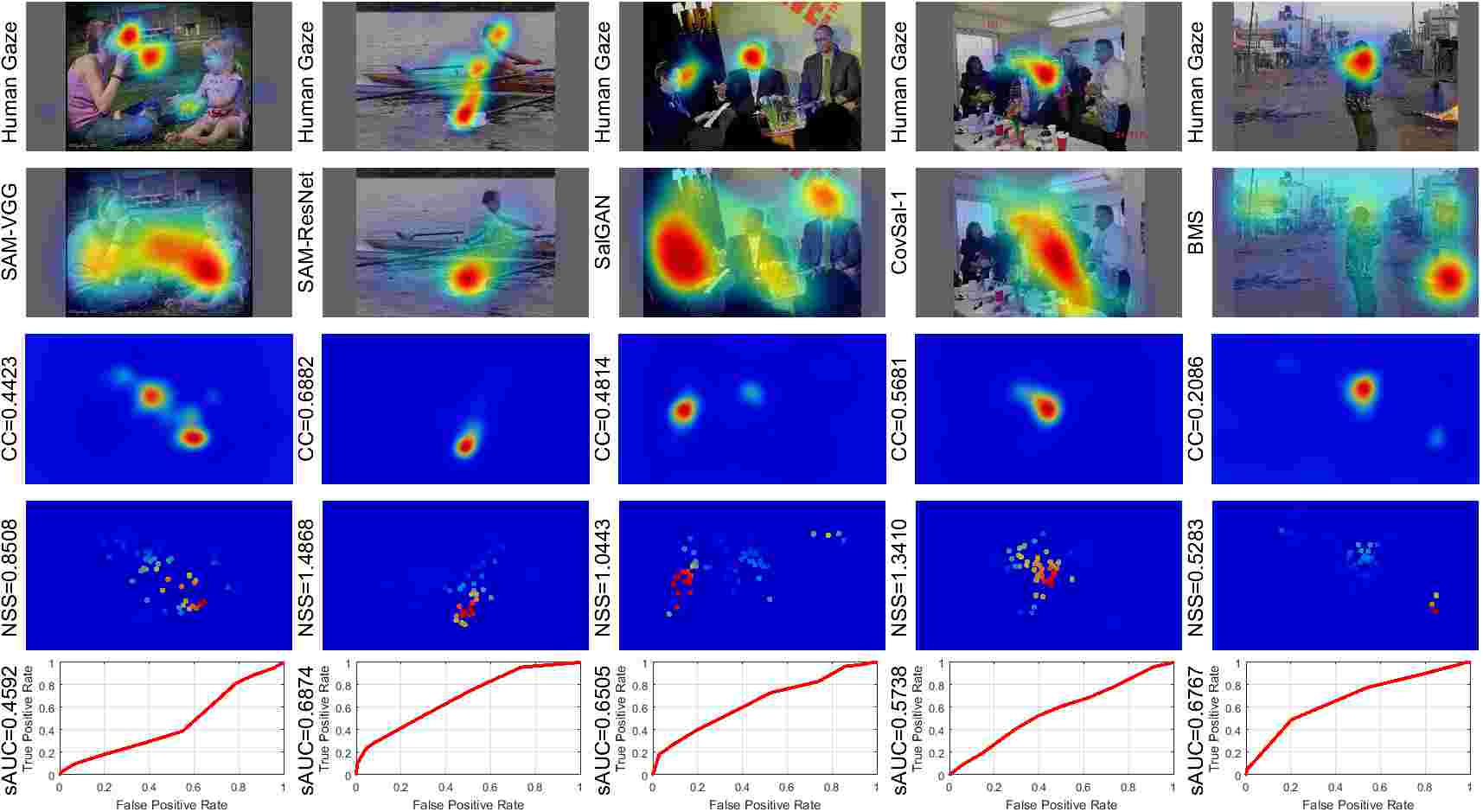}}\\
\vspace{-0.2cm}
\vspace{-0.1cm}
\caption{\small Example of failure cases of different models on Boundary and Rotation2. The 1st and 2nd rows of (b) are mapped by inverse transformation for better observation.}
\label{BR2}
\end{figure}

\textbf{Analyses of Metrics}: As shown in Figure \ref{ModelPer}, the ranks of saliency models and distortions are highly related to the evaluation metrics. Specifically, the Normalized Scanpath Saliency (NSS) is sensitive to false positives \cite{Bylinskii}.
sAUC, also called the shuffled AUC, penalizes models that include the center-bias and it ignores low-valued false positives compared to NSS \cite{Bylinskii}. As a result, the 1st and 5th columns in Figure \ref{BR2}.(a) have similar sAUC scores. NSS of the 5th column, however, is significantly lower than the 1st, because severe false positives on the 5th column contribute to lowering the normalized saliency value at each fixation location, thus reducing the overall NSS score. The same observation holds for the 3rd and 5th columns in Figure \ref{BR2}.(b).

\textbf{Outliers}: There are some outliers appearing on Figure \ref{ModelPer}, explained below.

\begin{figure}
\centering
\vspace{-0.1cm}
\tiny
\subfigure[\scriptsize Noise2]{\label{fig:edge-a}\includegraphics[height=0.26\linewidth]{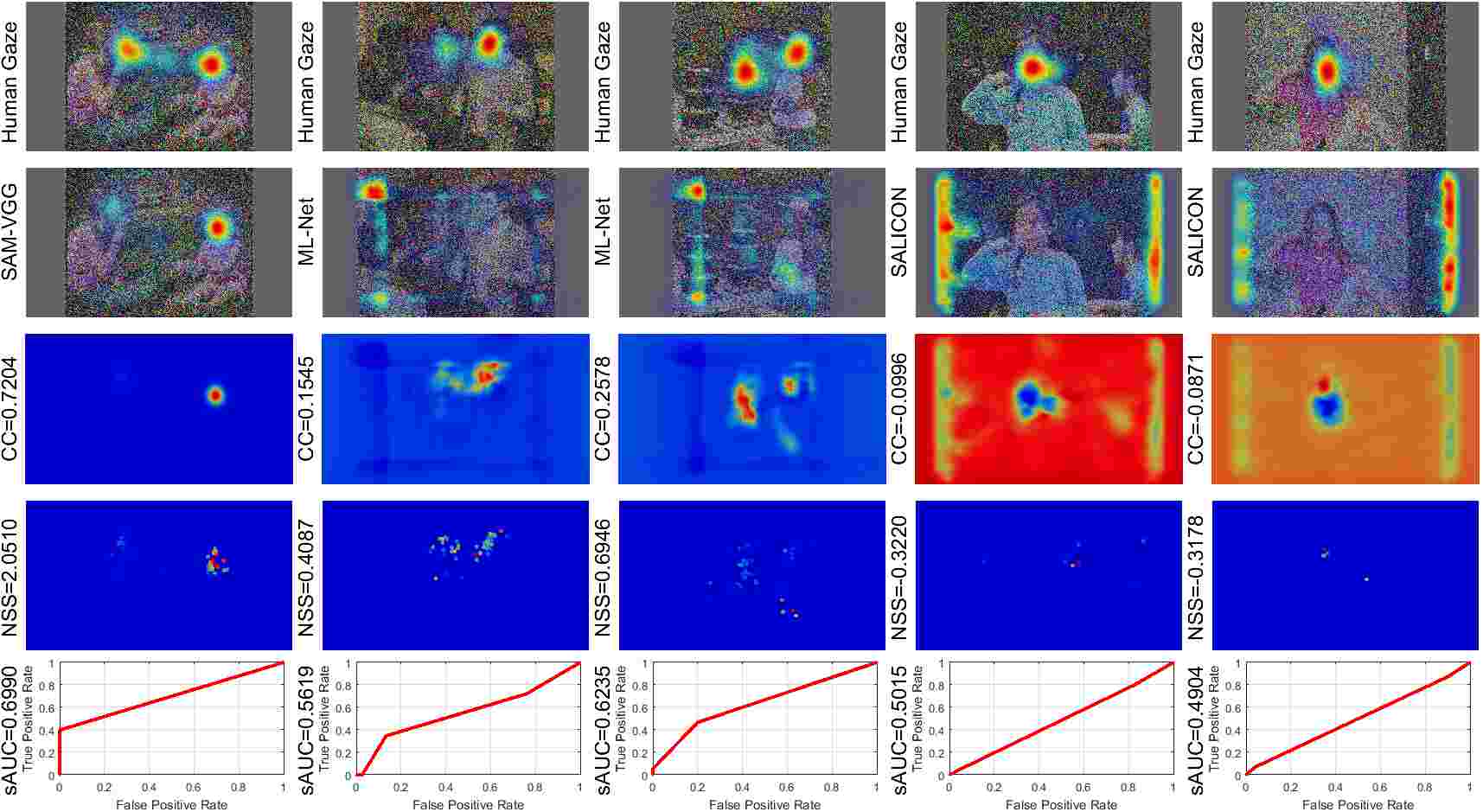}}\hspace{0.2cm}
\subfigure[\scriptsize Contrast2]{\label{fig:edge-a}\includegraphics[height=0.26\linewidth]{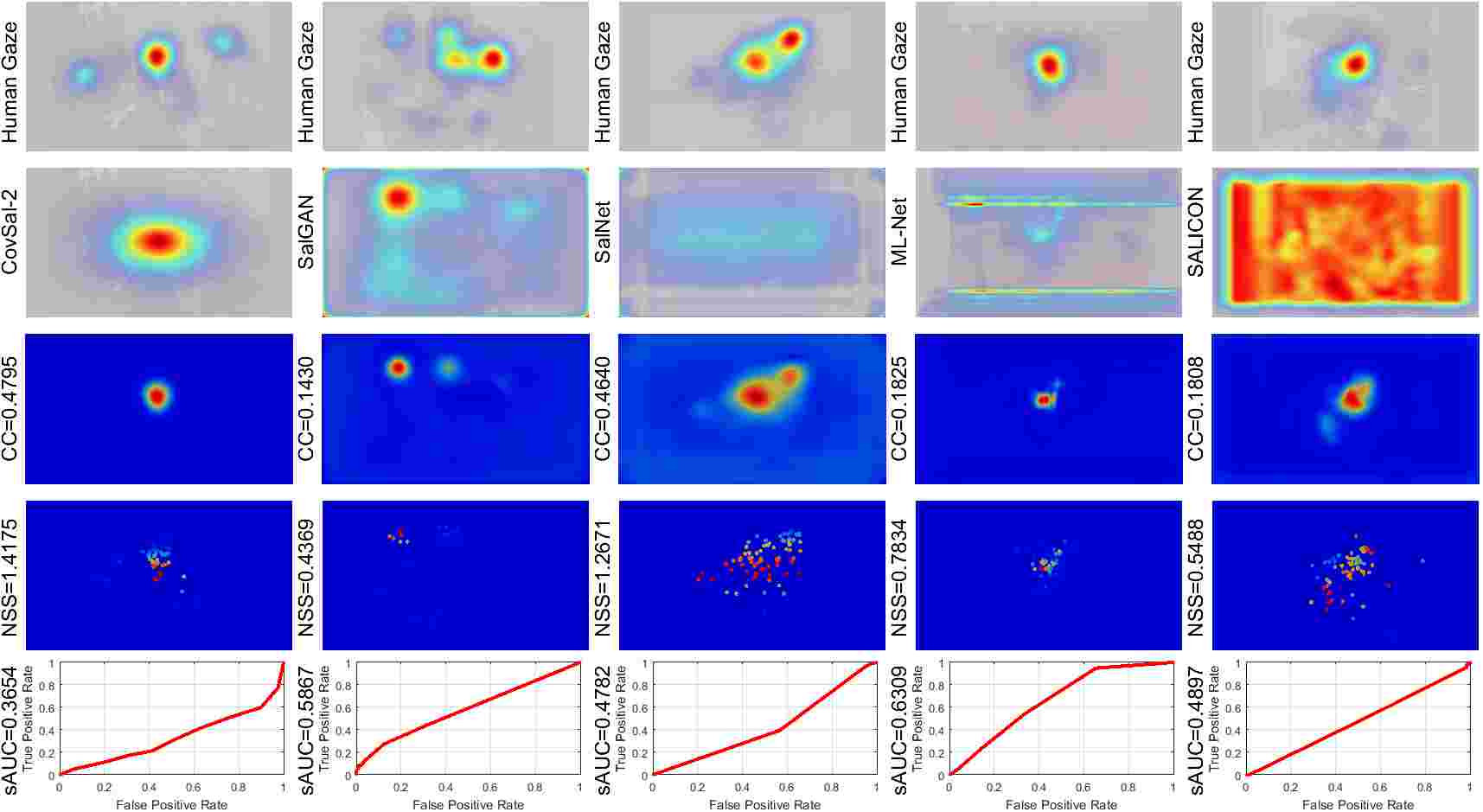}}\\
\vspace{-0.2cm}
\vspace{-0.1cm}
\caption{\small Example of failure cases of different models on Noise2 and Contrast2.}
\label{NC2}
\end{figure}

ML-Net fails on Noise2 and Contrast2. This is because ML-Net produces severe false positives on the upper-left region of stimulus corrupted by Noise2. Besides, ML-Net falsely produces two slender salient lines on the top and bottom sides of stimulus corrupted by Contrast2, as shown in Figure \ref{NC2}.

SALICON fails on Noise2 and Contrast2. Because SALICON produces severe false positives on the left and right sides of most stimulus corrupted by severe noise, as shown in Figure \ref{NC2}.(a). Further, SALICON detects the whole image as salient region on Contrast2 as shown in Figure \ref{NC2}.(b).

LS, LGS and GBVS fail on Boundary. LS extracts features from both of RGB and Lab color spaces, but the $a$ and $b$ channels of Lab color space are close to 0, because stimulus of Boundary are binary images. As a result, LS produces NaN values at a normalization step. Thus, we compute the LS saliency map
using only $L$ channel for avoiding NaN values. LGS and GBVS produce severe false positives on the left and right sides on Boundary as shown in Figure \ref{BR2}.(a).

CovSal-1 and CovSal-2 rank at the bottom on Figure \ref{ModelPer}.c, because CovSal includes severe center-bias which is penalized by the sAUC metric, as shown in Figure \ref{NC2}.(b).

\textbf{Upper-bound of models}: We report the Human Inter-Observer (IO) scores \cite{LSLGS} of different distortions in Table \ref{tab:DisGrou}. IO score provides an upper-bound on prediction accuracy of saliency models, because different observers are often the best predictors of each other.

\begin{figure}
\centering
\vspace{-0.1cm}
\tiny
\subfigure[\scriptsize MotionBlur2]{\label{fig:edge-a}\includegraphics[height=0.26\linewidth]{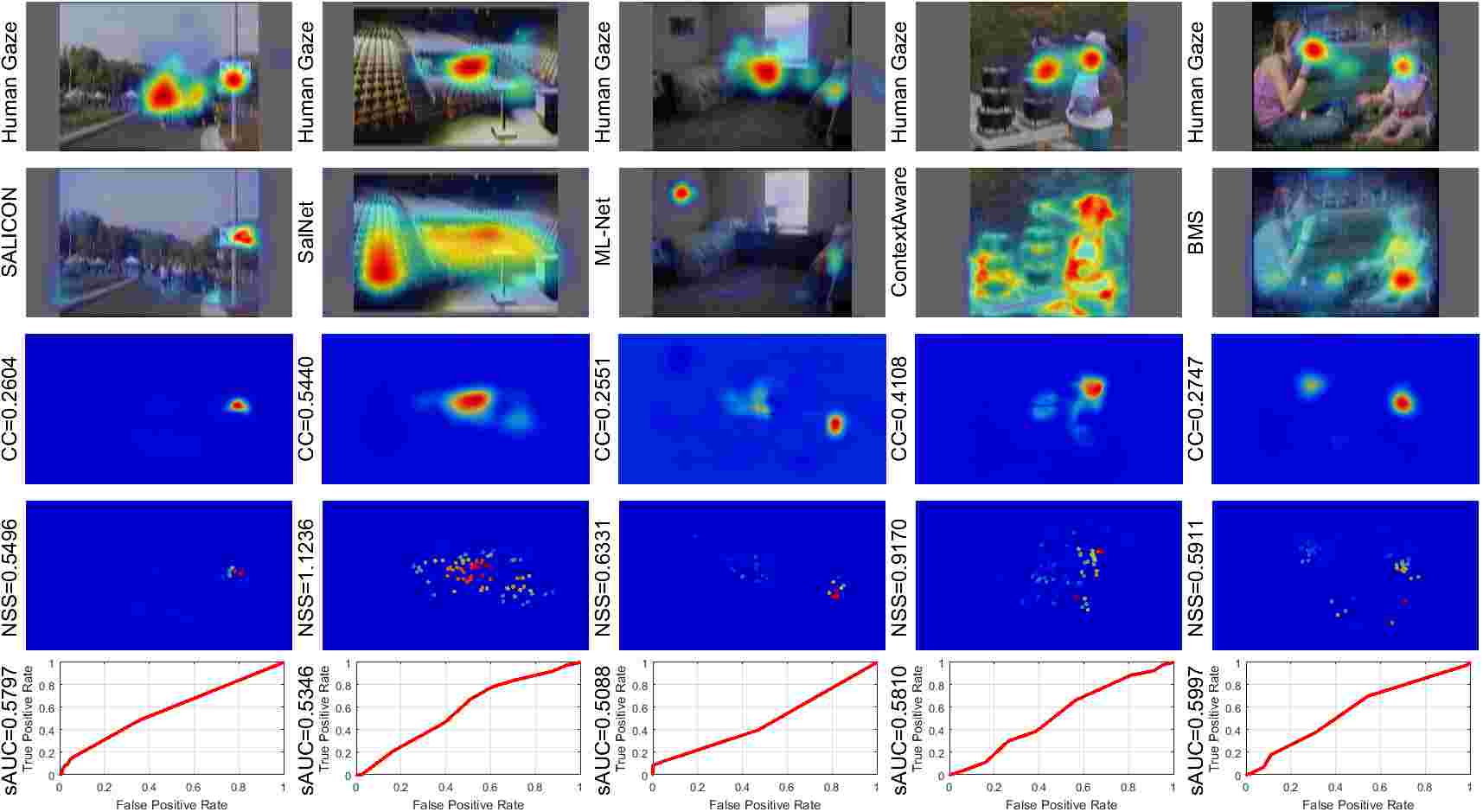}}\hspace{0.2cm}
\subfigure[\scriptsize Shearing3]{\label{fig:edge-a}\includegraphics[height=0.26\linewidth]{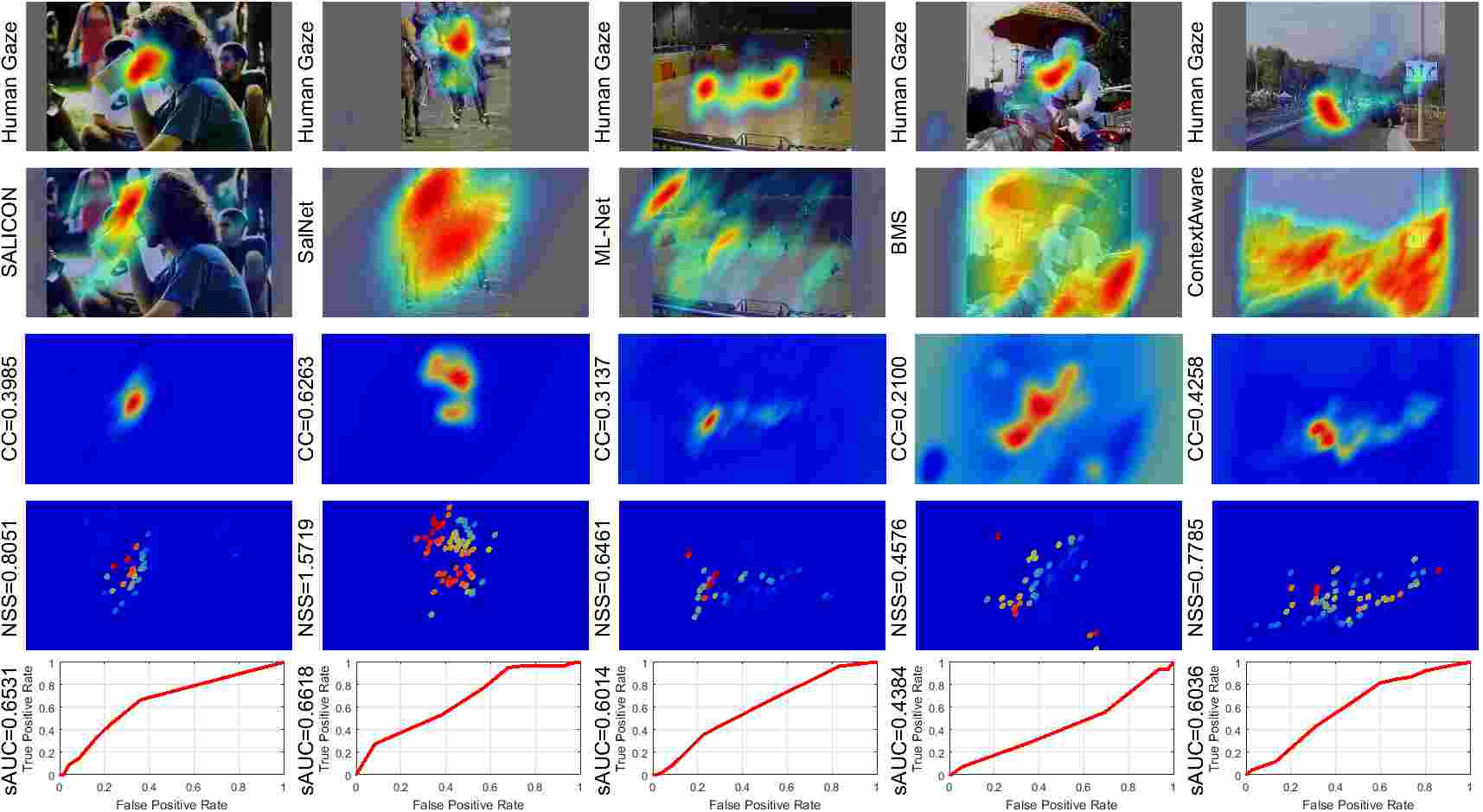}}\\
\vspace{-0.2cm}
\vspace{-0.1cm}
\caption{\small Example of failure cases of different models on MotionBlur2 and Shearing3. The 1st and 2nd rows of (b) are mapped by inverse transformation for better observation}
\label{M2S3}
\end{figure}

\textbf{Immunity to Distortion}: We define the \textbf{immunity} as the ability of a saliency model to predict the fixation locations for distorted stimuli as consistent as the distortion-free stimuli.
Figure \ref{ModelPer}.(d) shows the average sAUC score of different models on distorted stimuli when using human gaze on Reference as ground-truth. The higher score here means that the model has the better immunity to distortions, although it may not be the best model to predict the real human gaze when viewing distorted stimuli. We also find that most models obtain the higher sAUC scores when using human gaze on Reference as ground-truth compared to when using real human gaze distracted by distortions. Notably, the \textbf{immunity} mentioned here is different from the \textbf{robustness} mentioned in section 4.1. This is because the \textbf{robustness} means that the performance of saliency model will not be severely degraded by different distortions, and the performance is calculated by using real human gaze on distorted stimuli as ground-truth.

\section{Application for Data Augmentation}
Data augmentation is widely used in deep-learning based computer vision tasks \cite{DA1} to reduce overfitting and to improve generalization capacity of deep models. The most common data augmentation strategy is enlarging the dataset using some label-preserving transformations, such as Cropping, Inversion, ContrastChange, and Rotation. However, different from classical image classification and object detection problems, common data augmentation methods may produce label noise for the saliency prediction problem, because different transformations will change the ground truth at different levels. This paper carries important implications as to which of these kinds of transformations are valid and which are not as approximations of real human gaze behavior. We divide common transformations of the proposed dataset into two sets, i.e. valid and invalid augmented sets, and explore how fine-tuning on different sets of augmented data can improve or degrade the performance of deep models with respect to ground truth data.

\begin{table}
\centering
\renewcommand{\arraystretch}{1.6}
\scriptsize
    \caption{\small The performance of deep models on valid augmented set. Metric scores are represented by different colors: \textcolor{red}{sAUC}, \textcolor{green}{CC}, \textcolor{blue}{NSS}, \textcolor{cyan}{SIM}, \textcolor{magenta}{AUC-Borji}, and \textcolor{black}{KL}.}
    \vspace{-10pt}
    \label{tab:FineTune1}
    \tiny
    \centering
    \begin{tabular}{|c|c|c|c|c|}
    \hline
    {\bfseries } &{\bfseries SAM-VGG} &{\bfseries SAM-ResNet } &{\bfseries ML-Net } &{\bfseries SALICON } \\
   \hline
   \hline
    \tabincell{c}{\bfseries Original \\ \bfseries Model} &\tabincell{c}{\textcolor{red}{0.6330}, \textcolor{green}{0.4795}, \textcolor{blue}{1.5697}, \\ \textcolor{cyan}{0.4639}, \textcolor{magenta}{0.7434}, \textcolor{black}{0.9975}}  &\tabincell{c}{\textcolor{red}{0.6568}, \textcolor{green}{0.5304}, \textcolor{blue}{1.6716}, \\ \textcolor{cyan}{0.4817}, \textcolor{magenta}{0.8132}, \textcolor{black}{0.8776}.} &\tabincell{c}{\textcolor{red}{0.6588}, \textcolor{green}{0.5887}, \textcolor{blue}{1.9443}, \\ \textcolor{cyan}{0.5146}, \textcolor{magenta}{0.7930}, \textcolor{black}{0.8834}.} &\tabincell{c}{\textcolor{red}{0.6548}, \textcolor{green}{0.4774}, \textcolor{blue}{1.5969}, \\ \textcolor{cyan}{0.4561}, \textcolor{magenta}{0.7653}, \textcolor{black}{0.9826}.}\\
    \hline
    \tabincell{c}{\bfseries Fine-tuned \\ \bfseries using CAT2000} &\tabincell{c}{\textcolor{red}{0.6386}, \textcolor{green}{0.7606}, \textcolor{blue}{2.2692}, \\ \textcolor{cyan}{0.6317}, \textcolor{magenta}{0.8681}, \textcolor{black}{1.0122}}  &\tabincell{c}{\textcolor{red}{0.6546}, \textcolor{green}{0.7730}, \textcolor{blue}{2.2922}, \\ \textcolor{cyan}{0.6385}, \textcolor{magenta}{0.8724}, \textcolor{black}{1.1512}.} &\tabincell{c}{\textcolor{red}{0.6439}, \textcolor{green}{0.5974}, \textcolor{blue}{1.9558}, \\ \textcolor{cyan}{0.5364}, \textcolor{magenta}{0.8107}, \textcolor{black}{0.8945}.} &\tabincell{c}{\textcolor{red}{0.6588}, \textcolor{green}{0.5304}, \textcolor{blue}{1.6223}, \\ \textcolor{cyan}{0.4778}, \textcolor{magenta}{0.8205}, \textcolor{black}{0.8691}.}\\
    \hline
     \tabincell{c}{\bfseries Fine-tuned  \\ \bfseries using Valid Set}  &\tabincell{c}{\textcolor{red}{0.6442}, \textcolor{green}{0.7753}, \textcolor{blue}{2.3444}, \\ \textcolor{cyan}{0.6584}, \textcolor{magenta}{0.8766}, \textcolor{black}{0.6673}.}   &\tabincell{c}{\textcolor{red}{0.6667}, \textcolor{green}{0.7822}, \textcolor{blue}{2.3567}, \\ \textcolor{cyan}{0.6627}, \textcolor{magenta}{0.8817}, \textcolor{black}{0.6821}.} &\tabincell{c}{\textcolor{red}{0.6614}, \textcolor{green}{0.5984}, \textcolor{blue}{1.9527}, \\ \textcolor{cyan}{0.5386}, \textcolor{magenta}{0.8207}, \textcolor{black}{0.7702}.} &\tabincell{c}{\textcolor{red}{0.6598}, \textcolor{green}{0.5484}, \textcolor{blue}{1.7108}, \\ \textcolor{cyan}{0.5018}, \textcolor{magenta}{0.8469}, \textcolor{black}{0.8134}.} \\

     \hline
    \end{tabular}
\vspace{-10pt}
\end{table}

\begin{table}
\centering
\renewcommand{\arraystretch}{1.6}
\scriptsize
    \caption{\small The performance of deep models on invalid augmented set. }
    \vspace{-10pt}
    \label{tab:FineTune2}
    \tiny
    \centering
    \begin{tabular}{|c|c|c|c|c|}
    \hline
    {\bfseries } &{\bfseries SAM-VGG} &{\bfseries SAM-ResNet } &{\bfseries ML-Net } &{\bfseries SALICON } \\
   \hline
   \hline
    \tabincell{c}{\bfseries Original \\ \bfseries Model} &\tabincell{c}{\textcolor{red}{0.5989}, \textcolor{green}{0.5026}, \textcolor{blue}{1.4154}, \\ \textcolor{cyan}{0.4961}, \textcolor{magenta}{0.7226}, \textcolor{black}{1.0108}}  &\tabincell{c}{\textcolor{red}{0.5990}, \textcolor{green}{0.5026}, \textcolor{blue}{1.4154}, \\ \textcolor{cyan}{0.4961}, \textcolor{magenta}{0.7224}, \textcolor{black}{1.0168}.} &\tabincell{c}{\textcolor{red}{0.5938}, \textcolor{green}{0.5552}, \textcolor{blue}{1.4762}, \\ \textcolor{cyan}{0.5230}, \textcolor{magenta}{0.7450}, \textcolor{black}{0.9174}.} &\tabincell{c}{\textcolor{red}{0.5943}, \textcolor{green}{0.4179}, \textcolor{blue}{1.1162}, \\ \textcolor{cyan}{0.4535}, \textcolor{magenta}{0.7154}, \textcolor{black}{0.9833}.}\\
    \hline
    \tabincell{c}{\bfseries Fine-tuned \\ \bfseries using CAT2000} &\tabincell{c}{\textcolor{red}{0.5790}, \textcolor{green}{0.7551}, \textcolor{blue}{1.9231}, \\ \textcolor{cyan}{0.6507}, \textcolor{magenta}{0.8406}, \textcolor{black}{0.9949}}  &\tabincell{c}{\textcolor{red}{0.5771}, \textcolor{green}{0.7584}, \textcolor{blue}{1.8984}, \\ \textcolor{cyan}{0.6517}, \textcolor{magenta}{0.8423}, \textcolor{black}{1.0745}.} &\tabincell{c}{\textcolor{red}{0.5845}, \textcolor{green}{0.5727}, \textcolor{blue}{1.4433}, \\ \textcolor{cyan}{0.5442}, \textcolor{magenta}{0.7652}, \textcolor{black}{0.7949}.} &\tabincell{c}{\textcolor{red}{0.6080}, \textcolor{green}{0.5379}, \textcolor{blue}{1.3662}, \\ \textcolor{cyan}{0.5194}, \textcolor{magenta}{0.7978}, \textcolor{black}{0.7793}.}\\
    \hline
     \tabincell{c}{\bfseries Fine-tuned  \\ \bfseries using Invalid Set}  &\tabincell{c}{\textcolor{red}{0.5751}, \textcolor{green}{0.7268}, \textcolor{blue}{1.8291}, \\ \textcolor{cyan}{0.6316}, \textcolor{magenta}{0.8233}, \textcolor{black}{1.6512}.}   &\tabincell{c}{\textcolor{red}{0.5716}, \textcolor{green}{0.7508}, \textcolor{blue}{1.8880}, \\ \textcolor{cyan}{0.6490}, \textcolor{magenta}{0.8263}, \textcolor{black}{1.7643}.} &\tabincell{c}{\textcolor{red}{0.5827}, \textcolor{green}{0.5546}, \textcolor{blue}{1.3898}, \\ \textcolor{cyan}{0.5402}. \textcolor{magenta}{0.7594}, \textcolor{black}{1.1768}.} &\tabincell{c}{\textcolor{red}{0.6017}, \textcolor{green}{0.5341}, \textcolor{blue}{1.3434}, \\ \textcolor{cyan}{0.5065}, \textcolor{magenta}{0.7918}, \textcolor{black}{0.7919}.} \\

     \hline
    \end{tabular}
\vspace{-15pt}
\end{table}

On the one hand, we select Reference, Mirroring, Inversion, Contrast1, Shearing1, JPEG1 and Noise1 to generate a valid augmented set, because these transformations have slight effects on human gaze. On the other hand, Rotation1, Rotation2, Shearing2, Shearing3, Cropping1, Cropping3 and MotionBlur2 serve as an invalid augmented set, because these transformations are not able to preserve human gaze labels as approximations of Reference. Considering that Reference images of the proposed dataset is selected from CAT2000, here we first fine-tune 4 state-of-the-art deep models using CAT2000. Then, we use the valid and invalid augmented sets to fine-tune these deep models separately.

For fair comparison, we unify the training set scale, optimization function parameters, and training epoch for different fine-tuning strategies using CAT2000, valid and invalid sets, explained below. First, each of valid set, invalid set and CAT2000 used here is divided into training set (350 images), validation set (175 images), and test set (175 images). Second, we adopt the test set of valid augmented set to calculate the performance in Table \ref{tab:FineTune1}, and the test set of invalid augmented set is used to compute metric scores in Table \ref{tab:FineTune2}. Third, for each deep model, the hyper-parameters of fine-tuning this model by CAT2000, valid set, and invalid set are set as the same values: 1) For 4 deep models mentioned in Table \ref{tab:FineTune1}, SGD (stochastic gradient descent) with momentum 0.9 and weight decay 0.0005 serves as the optimization function, and the batch size is 1, 2) For ML-Net, learning rate is $\rm 10^{-2}$, and epoch is 20, 3) For SALICON, learning rate is $\rm 10^{-6}$, and training time is set as 2000 seconds, and 4) For SAM-VGG and SAM-ResNet, learning rates are set as $\rm 3\times10^{-7}$, and epoches are 10.

\begin{figure}
\vspace{-0.1cm}
\centering
\vspace{-0.1cm}
\subfigure[\scriptsize SAM-VGG]{\label{fig:edge-a}\includegraphics[height=0.22\linewidth]{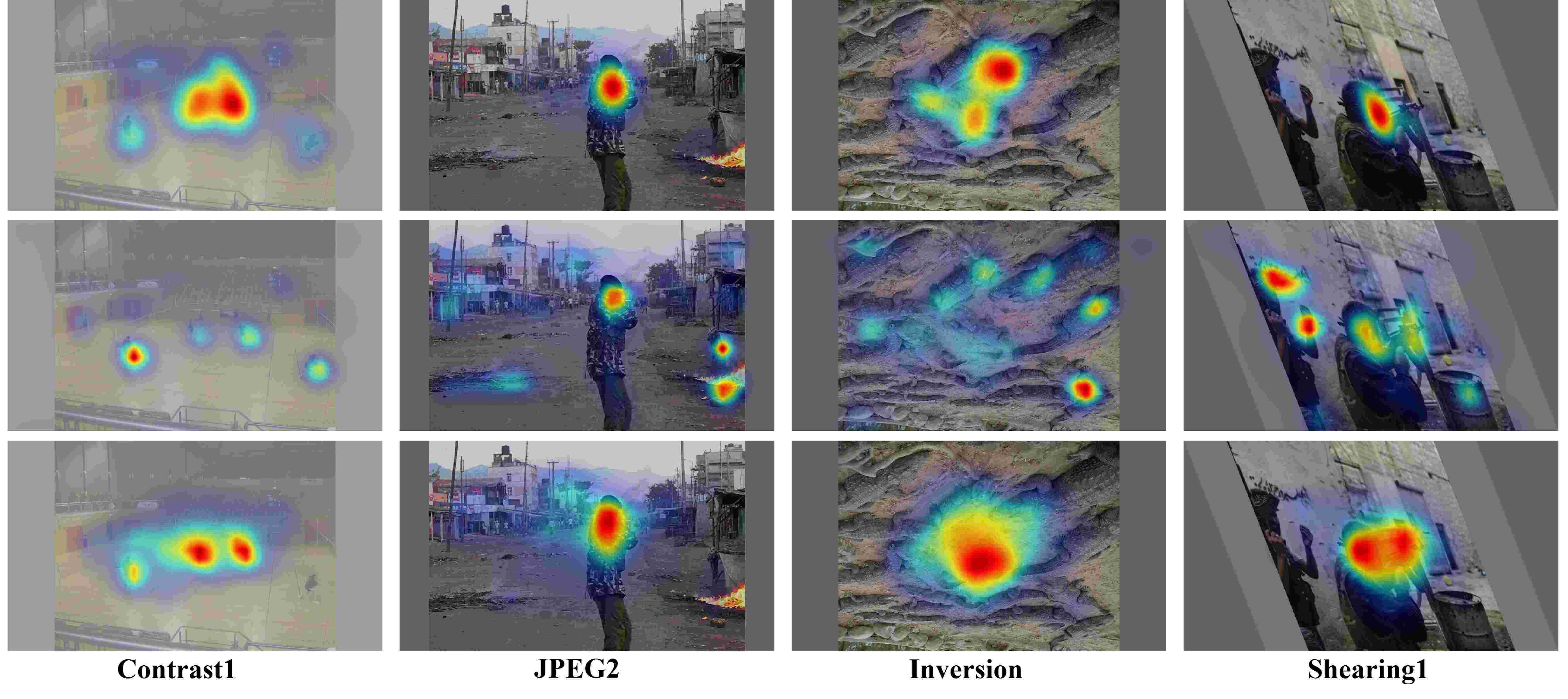}}
\subfigure[\scriptsize SALICON]{\label{fig:edge-a}\includegraphics[height=0.22\linewidth]{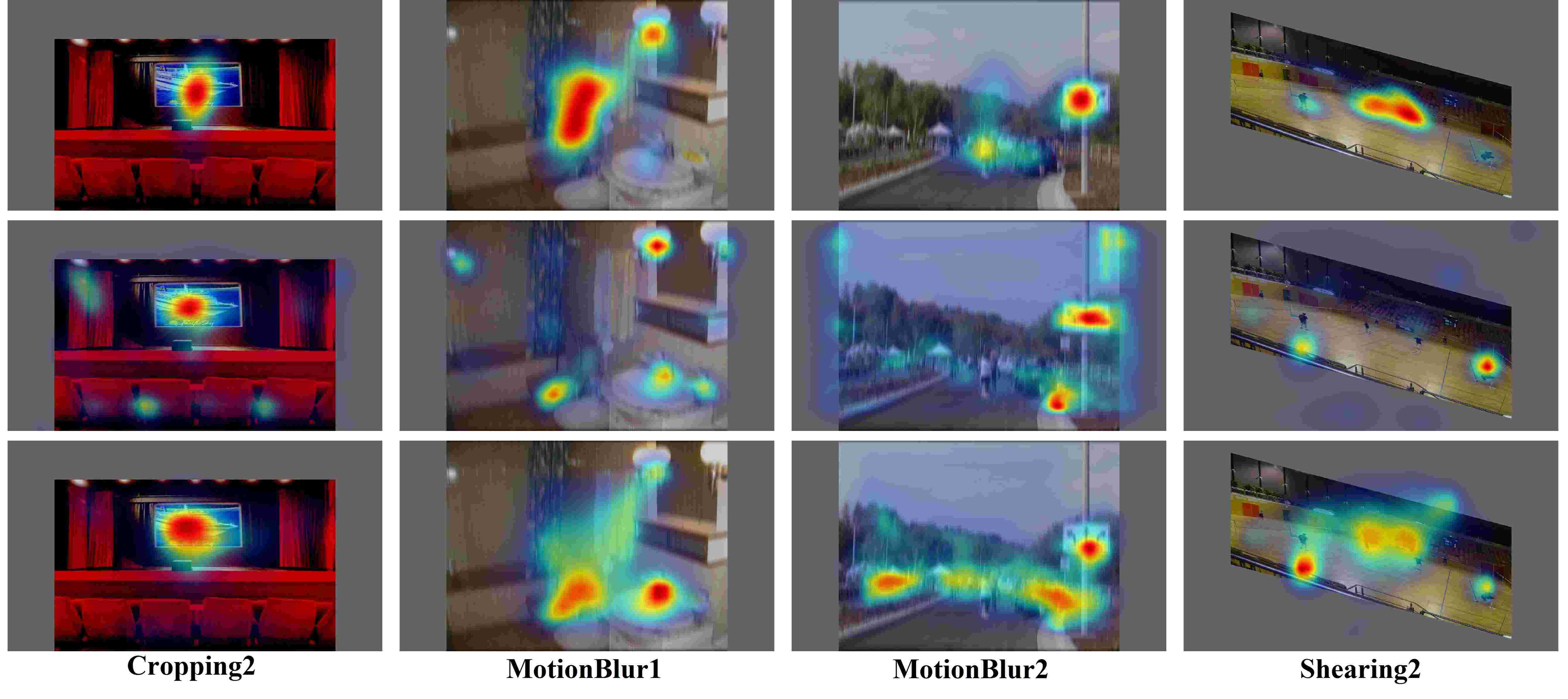}}\\
\subfigure[\scriptsize ML-Net]{\label{fig:edge-a}\includegraphics[height=0.22\linewidth]{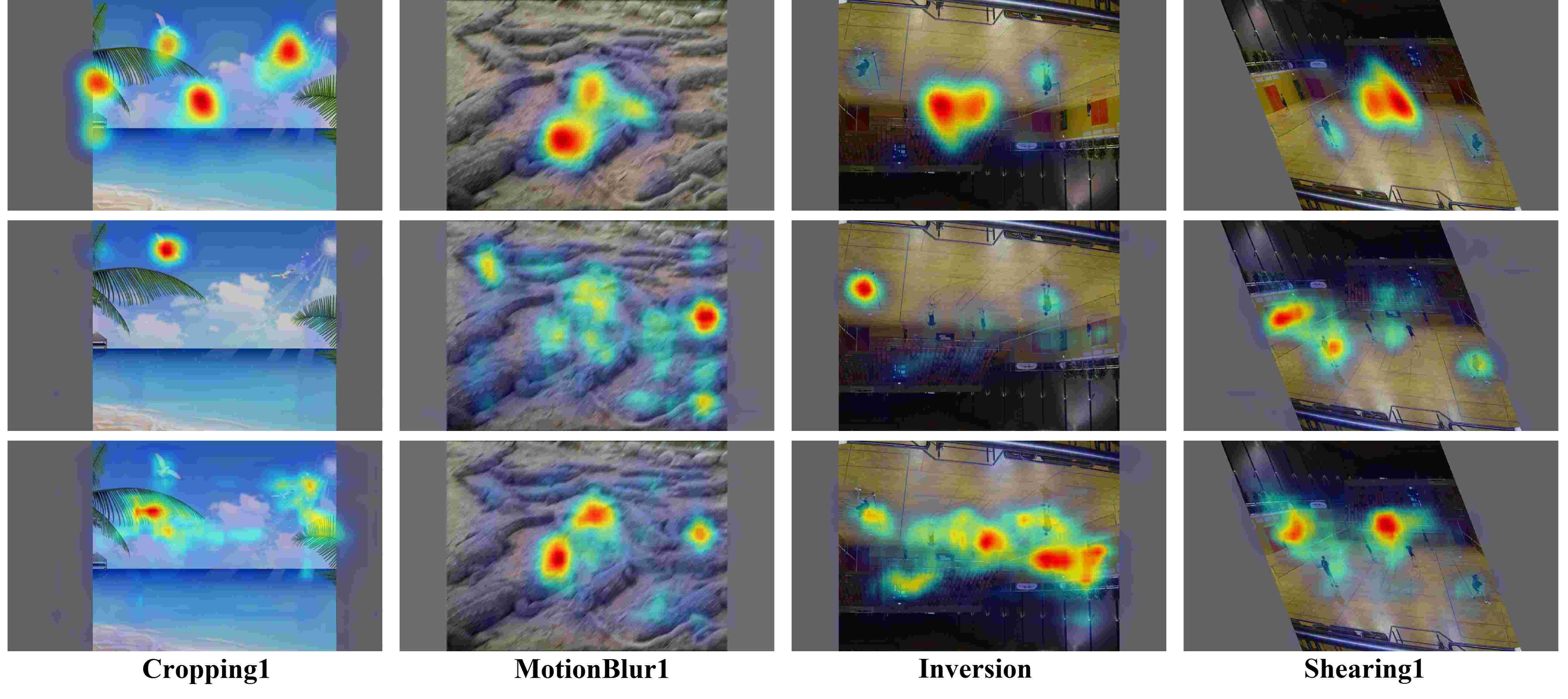}}
\subfigure[\scriptsize SAM-ResNet]{\label{fig:edge-a}\includegraphics[height=0.22\linewidth]{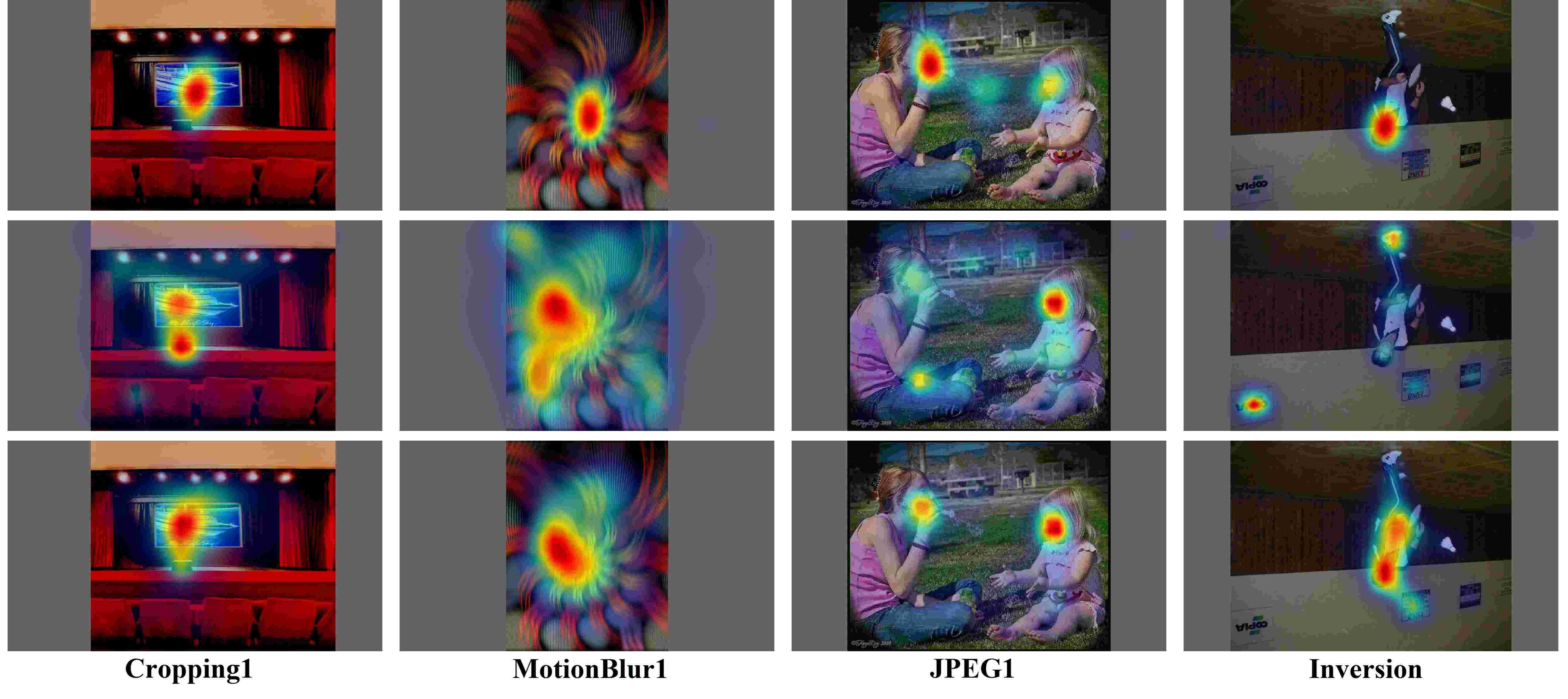}}
\vspace{-0.2cm}
\vspace{-0.1cm}
\caption{\small Qualitative comparison between ground truth, original models, and fine-tuned models. For (a)-(d), the 1st row represents ground truth of human gaze; the 2nd row represents saliency maps generated by original models; the 3rd row represents saliency maps of refined models fine-tuned by valid set.}
\label{bigfinal}
\end{figure}

Experimental results shown in Table \ref{tab:FineTune1} verify that fine-tuning using CAT2000 and valid set can improve deep models' performance. Besides, fine-tuning using valid set achieves better promotion compared to using CAT2000 which contains only normal Reference images. However, as shown in Table \ref{tab:FineTune2}, fine-tuning using invalid set degrades deep models' performance compared to using CAT2000. Qualitative results generated by original models and refined models which are fine-tuned using valid set, together with ground truth, are shown in Figure \ref{bigfinal}. From a qualitative point of view, for the saliency prediction task, fine-tuning using some label-preserving data augmentation transformations (as mentioned in valid set) can improve robustness of deep saliency models against distortions.

\section{Discussion and Conclusion}
We introduce a large-scale eye-tracking database consisting of 19 typical distortions for boosting saliency modeling to approach human-level accuracy on non-canonical stimuli. We also refine some state-of-the-art deep saliency models by valid data augmentation strategy to achieve better prediction accuracy when suffering from distortions.

The takeaway lessons from our study are as follows.

First, most distortions do have impacts on human gaze, and the magnitude of impact highly depends on distortion type. High level rotation, shearing and cropping distortions significantly distract human gaze. While mirroring, inversion and slight shearing distortions have slight impacts on human gaze.

Second, different distortions distract human gaze in different ways. For example, extreme-low contrast attracts human gaze to center region. Rotation alters the dominant salient object. Cropping distracts human gaze from the salient region appearing on the cut side.


Third, deep saliency models obtain better average performance on different distortions than early non-deep models, but fail on some special distortions including severe noise, boundary and extreme-low contrast. Early saliency models using hand-crafted features provide better robustness in these cases.

Finally, for saliency prediction problem, how to choose data augmentation transformation types has impact on final performance of deep saliency models. Mirroring, Inversion, Contrast1, Shearing1, JPEG1 and Noise1 are qualified to serve as data augmentation transformations, and to improve model performance. While Cropping, Rotation, Shearing2, Shearing3, MotionBlur2 will degrade model performance because these transformations change human gaze label severely.

For state-of-the-art saliency models, there is still a gap between the current prediction and the upper-bound (IO score) of prediction accuracy on distorted stimuli.

We will share our collected data and code with the community to promote research in improving the robustness of deep models over different distortions and to close the gap between saliency models and the human IO model.


\bibliographystyle{splncs}
\bibliography{egbib}

\end{document}